\documentclass[sn-mathphys-num]{sn-jnl}

\usepackage{multirow}%
\usepackage{amsmath,amssymb,amsfonts}%
\usepackage{amsthm}%
\usepackage{mathrsfs}%
\usepackage[title]{appendix}%
\usepackage{xcolor}%
\usepackage{textcomp}%
\usepackage{manyfoot}%
\usepackage{booktabs}%

\usepackage{graphicx}%
\definecolor{linkgreen}{HTML}{01a049}
\definecolor{linkblack}{HTML}{0000EE}
\usepackage{tikz,pgfplots}
\usepackage{lipsum} 
\usepackage{xcolor}
\usepackage{algorithmic}
\usepackage{algorithm}
\usepackage{subcaption}
\usepackage{enumitem}
\usepackage{array}
\newcommand{\mG}{\mathbf{G}}
\newcommand{\mD}{\mathbf{D}}

\newcommand{\mC}{\mathbf{C}}
\newcommand{\mQ}{\mathbf{Q}}
\newcommand{\mF}{\mathbf{F}}

\newcommand{\mA}{\mathbf{A}}

\newcommand{\mU}{\mathbf{U}}
\newcommand{\mX}{\mathbf{X}}
\newcommand{\mW}{\mathbf{W}}
\newcommand{\mI}{\mathbf{I}}

\newcommand{\mB}{\mathbf{B}}

\newtheorem{proper}{proper}

\newsavebox\CBox
\def\textBF#1{\sbox\CBox{#1}\resizebox{\wd\CBox}{\ht\CBox}{\textbf{#1}}}
\usepackage{float}

\begin{document}

\title[Scalable Multi-view Clustering via Explicit Kernel Features Maps]{Scalable Multi-view Clustering via Explicit Kernel Features Maps}


\author*[1,2]{\fnm{Chakib} \sur{Fettal}}\email{chakib.fettal@u-paris.fr}

\author[1]{\fnm{Lazhar} \sur{Labiod}}\email{lazhar.labiod@u-paris.fr}

\author[1]{\fnm{Mohamed} \sur{Nadif}}\email{mohamed.nadif@u-paris.fr}

\affil[1]{\orgdiv{Centre Borelli UMR 9010}, \orgname{Université Paris Cité}}
\affil[2]{\orgname{CDC Informatique}}


\abstract{
The proliferation of high-dimensional data from sources such as social media, sensor networks, and online platforms has created new challenges for clustering algorithms. Multi-view clustering, which integrates complementary information from multiple data perspectives, has emerged as a powerful solution. However, existing methods often struggle with scalability and efficiency, particularly on large attributed networks. In this work, we address these limitations by leveraging explicit kernel feature maps and a non-iterative optimization strategy, enabling efficient and accurate clustering on datasets with millions of points. 
}



\maketitle

\begin{figure}[t]
    \centering
    \includegraphics[width=\textwidth]{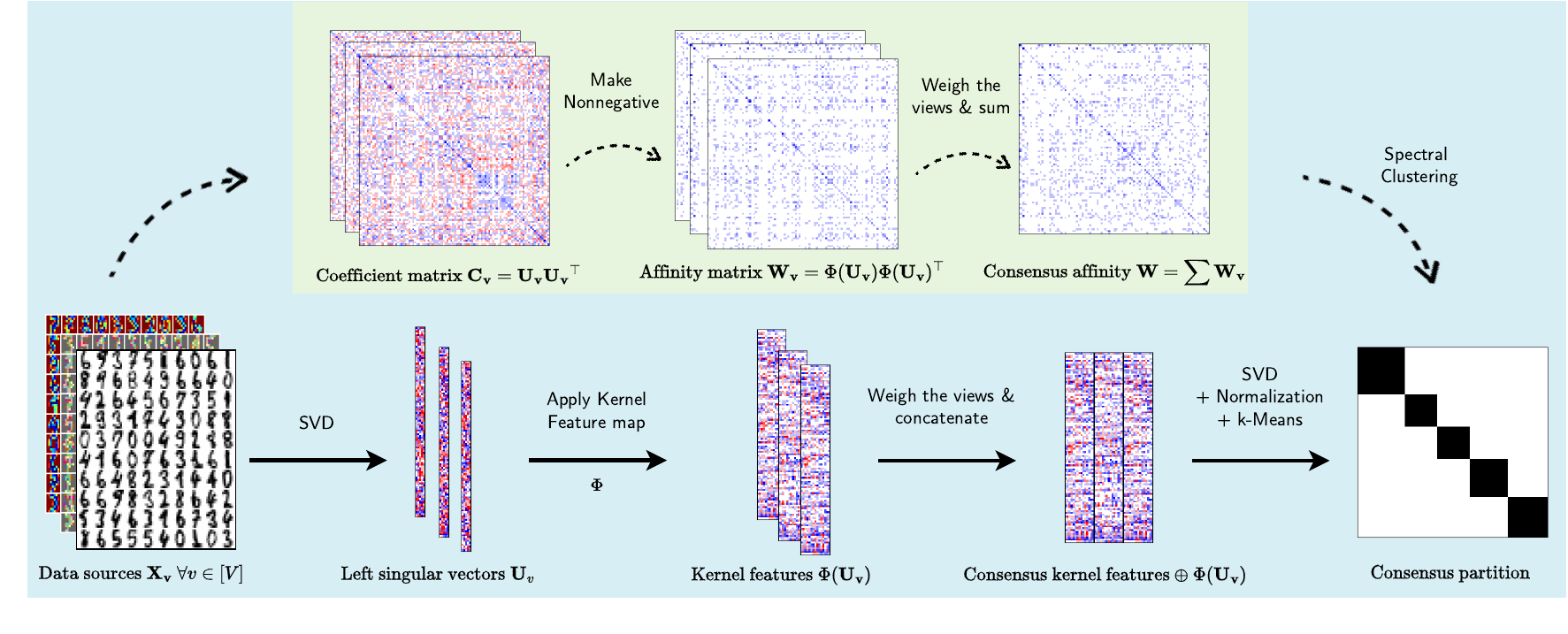}
    \caption{Schematic diagram: the black section represents the operations that are explicitly performed in our approach; the green section represents what is transpiring implicitly (operations in the same column represent the same step). Given a set of inputs, we apply an singular value decomposition (SVD) to obtain the left singular values (equivalent to implicitly computing a coefficient matrix using low-rank subspace clustering). We then apply a nonnegative feature map (equivalent to implicitly computing an affinity matrix). We then weigh and concatenate the obtained embeddings of each view (equivalent to summing the affinity matrices into a single consensus one). Finally, we perform an SVD followed by a k-Means to obtain a consensus partition (equivalent to performing spectral clustering on the consensus affinity matrix). \textcolor{black}{Novelty comes from extending \cite{fettal2023scalable} to a multi-view setting in an efficient manner through the properties of kernels.}}
    \label{fig:schema}
\end{figure}

\section{Introduction}
The explosive growth of data from diverse sources, such as social media, sensor networks, and online platforms, has led to the emergence of complex high-dimensional datasets. These datasets often contain multiple views of the same underlying data, each capturing different aspects or perspectives. Traditional clustering algorithms, designed for single-view data, face significant challenges in effectively capturing intricate relationships and structures that extend across multiple views. In this context, multi-view clustering, which integrates information from multiple views in order to enhance clustering performance, has emerged as a powerful paradigm. It has been successfully applied to various applications including computer vision \cite{yin2015multi,wang2016iterative}, natural language processing \cite{kumar2011co, liu2013multi, boutalbi2022tensor}, and bioinformatics \cite{sun2013multi,wang2016multi,sun2015multi,yu2020clustering}. For a more in-depth discussion, the reader can refer to \cite{chao2021survey, yu2025review}.

In this work, we are specifically interested in multiview subspace clustering \cite{gao2015multi,sun2021scalable,chen2022efficient,cai2023seeking,chen2025adaptive, zhong2025locality}. By exploring the latent subspaces within each view and leveraging the complementary nature of different views, multi-view subspace clustering algorithms can reveal hidden structures that may not be apparent when considering individual views in isolation. This not only leads to more accurate clustering results, but also enables a deeper understanding of the underlying data. Specifically, multi-view clustering for attributed network data comprising interconnected entities with associated attributes offers a powerful framework for analyzing complex datasets and, for this reason, has gained significant attention in the recent literature \cite{fan2020one2multi, lin2021graph,lin2021multi,pan2021multi,liu2024beyond}. \textcolor{black}{One prominent application of multiview clustering for attributed network data is in social network analysis, since social networks often exhibit rich attributes relating to individuals or groups: demographic information, interests, affiliations, etc. \cite{fan2020one2multi,hu2020open}. Another application is in biological research, where multi-omics data can be integrated using multiple graph convolutional networks \cite{wang2021mogonet}. Finally, recommender systems often rely on multiple sources of information, such as user preferences, item attributes, and social connections \cite{kipf2016semi,wu2019simplifying,shchur2018pitfalls}.}

The computational complexity of multiview subspace clustering has hindered its widespread adoption in real-world applications, although some efficient approaches have been proposed. These can involve, for example, anchor-based techniques to reduce the size of matrices used in optimization \cite{kang2020large,sun2021scalable}. \textcolor{black}{However, our experiments show that existing methods often struggle with efficiency on large-scale multi-view datasets, and therefore, further computational improvements are needed.}

\textcolor{black}{To address the scalability issue of multiview subspace clustering, rather than using anchor-based methods, we decided to take advantage of the properties of kernel feature maps. Indeed, the class of random features is a popular technique for accelerating kernel methods in large-scale problems, recognized by prestigious awards. \cite{rahimi2007random} proposed mapping the the input data to a randomized low-dimensional feature space and applying fast linear methods. \cite{cho2009kernel} introduced positive-definite kernel functions mimicking large neural nets, usable in both shallow and deep architectures. \cite{maji2009max} and \cite{vedaldi2012efficient} developed explicit feature maps for various kernels, significantly reducing train/test times. \cite{pham2013fast} introduced Tensor Sketching, a faster technique to approximate polynomial kernels, and \cite{liu2021random} provided a comprehensive survey on random features, summarizing algorithms, theoretical results and their relationship with deep neural networks. It is worth noting that none of these approaches have been harnessed in the context of clustering, and even less so in the context of multiview clustering, which is the focus of this paper. Our main contributions are:}

\begin{enumerate}
\item First, we introduce a novel framework that integrates multiple views of the data subspace structure graph into a unified consensus subspace structure graph, effectively capturing the complementary information across views and allowing for a new scalability framework for multi-view subspace clustering. 
\item Second, we propose an efficient optimization strategy that scales well with large graphs. Using the properties of kernel feature maps, our algorithm significantly reduces computational burden while maintaining state-of-the-art clustering performance. This scalability makes our approach suitable for handling real-world applications, since it is able to scale to datasets with millions of data points.
\item Third, we conduct extensive experiments, including statistical significance testing, on real-world benchmark networks of different scales and \textcolor{black}{degrees of overlap,} to evaluate our algorithm against state-of-the-art multi-view subspace clustering methods and attributed-network-specific multi-view approaches. Our experiments are entirely reproducible, ensuring transparency and enabling others to validate our results. The entirety of our source code is provided \footnote{\url{https://github.com/chakib401/MvSCK}}.
\end{enumerate}
The remainder of this paper is structured as follows. In Section 2 we discuss related works. Section 3 introduces the necessary background to develop our approach. In Section 4 we develop our proposed method and explore the algorithm and its complexity. Section 5 contains a detailed description of our experiments. Finally, we give our conclusion in Section 6. A schematic representation of our approach is shown in Figure \ref{fig:schema}. This paper extends our previous work \cite{fettal2023scalable}, where we introduced a scalable attributed-graph subspace clustering method. Here, we generalize the approach to multi-view subspace clustering in a scalable manner by leveraging properties of kernel maps, and provide a more comprehensive experimental evaluation on large-scale network datasets.

\section{Related Work}
This section discusses prior work relating to scalable multi-view subspace clustering. We also look at graph multi-view clustering, since we use attributed networks to evaluate performance.

\subsection{Scalable Multi-view Subspace Clustering}
In \cite{lin2021multi} \textit{Large-Scale Multi-View Subspace Clustering in Linear Time} (\textBF{LMVSC}) \cite{kang2020large} was introduced. This method, where smaller graphs are created for each view and then integrated to enable spectral clustering on a reduced graph size, derives from the concept of anchor graphs. The \textit{Scalable Multi-view Subspace Clustering with Unified Anchors} (\textBF{SMVSC}) \cite{sun2021scalable} approach integrates anchor learning and graph construction into a unified optimization framework, resulting in a more accurate representation of latent data distribution by learned anchors and a more discriminative clustering structure. In \cite{chen2022efficient}, an efficient \textit{orthogonal multi-view subspace clustering} (\textBF{OMSC}) approach was proposed, featuring joint anchor learning, graph construction and partitioning. Recently, \cite{Zhang_Wang_Li_Zhang_Liu_Zhu_Liu_Zhou_Luo_2023} developed a \textit{Flexible and Diverse Anchor Graph Fusion for Scalable Multi-view Clustering} (\textBF{FDAGF}) that introduces a fusion strategy for multi-size anchor graphs.

\subsection{Graph Multi-view Clustering}
In \cite{pan2021multi}, a \textit{multi-view contrastive graph clustering} (\textBF{MCGC}) method was proposed, designed to learn a consensus graph with contrastive regularization over the learned graph matrix. \cite{lin2021multi} introduced a novel \textit{multi-view attributed graph clustering} (\textBF{MAGC}) framework, which incorporates node attributes and graphs. It is an approach that stands out in three important respects: utilizing graph filtering for smooth node representation, rather than deep neural networks; learning a consensus graph from data to address noise and incompleteness in the original graph; and exploring high-order relations through a flexible regularizer design. The \textBF{MvAGC} model \cite{lin2021graph} is a simple yet effective approach for Multi-view Attributed Graph Clustering. It applies a graph filter to smooth the features without relying on neural network parameter learning. Reduction in computational complexity is achieved through a novel anchor point selection strategy. In addition, a new regularizer is introduced to explore high-order neighborhood information. \cite{fettal2023simultaneous} introduced \textBF{LMGEC}, a generic linear model for performing multi-view attributed graph representation learning and clustering simultaneously. It uses a simple weighing scheme based on inertia and single-hop neighborhood propagation.

\section{Motivation} Unlike existing multi-view subspace clustering methods that rely heavily on anchor-based graph reductions or iterative optimization, our approach—MvSCK—achieves scalability through the use of explicit kernel feature maps and kernel summation properties. This formulation enables efficient consensus affinity computation without constructing large intermediate graphs or solving costly iterative subproblems. Furthermore, while prior methods such as LMVSC, SMVSC, or FDAGF approximate scalability by learning reduced anchors, they still incur significant memory and computational overhead as dataset size increases. In contrast, MvSCK factorizes the consensus kernel directly through low-dimensional feature embeddings, yielding linear complexity with respect to data size. This design allows our method to scale seamlessly to millions of nodes while maintaining competitive accuracy.

\section{Preliminaries}
In what follows, matrices are denoted in boldface uppercase letters and vectors are denoted in boldface lowercase letters. $[n]$ corresponds to the integer set $\{1,\ldots, n\}$. $\texttt{Tr}$ is the trace operator. $\oplus$ denotes column-wise concatenation. $\Vert.\Vert$ refers to the $\ell_2$ norm, unless otherwise specified.

\subsection{Subspace Clustering}
Given a data matrix $\mX \in \mathbb{R}^{n\times d}$ and a desired number of clusters $k$, most subspace clustering problems rely on the self-expression property, taking the following form: 
\begin{equation}
\begin{aligned}
\min_{\mC} \quad & \Vert \mX-\mC\mX\Vert^2 \,+\, \Omega(\mC) \quad \text{s.t.}\quad \mC\in \mathcal{C}
\end{aligned}
\end{equation}
where $\mC \in \mathbb{R}^{n \times n}$ is the coefficient matrix, $\Omega$ is a regularization function, and $\mathcal{C}$ is the set of feasible solutions.

In this context, we assume the data lies on a union of linear subspaces. A \textbf{linear subspace} $\mathcal{S} \subseteq \mathbb{R}^d$ is a set closed under linear combinations; geometrically, this represents a flat shape (like a line or a plane) passing through the origin. If a data point $\mathbf{x}_i$ belongs to subspace $\mathcal{S}$, it can be reconstructed as a linear combination of other points within $\mathcal{S}$.

Once a solution $\mC$ is obtained, a subspace affinity graph is constructed, represented by its adjacency matrix $\mW$. Typically, this is calculated as:
\begin{equation}
\mW = \frac{|\mC|+|\mC^\top|}{2}.
\end{equation}
Finally, a partition of the $k$ groups is obtained by spectral clustering on $\mW$. Different values of $\Omega$ and $\mathcal{C}$ correspond to different subspace clustering techniques. For example, it is possible to seek sparse solutions by using the $\ell_1$ norm:
\begin{equation}
\begin{aligned}
\min_{\mC} \quad & \Vert \mX-\mC\mX\Vert^2 \,+\, \lambda \Vert\mC\Vert_1 \quad \text{s.t.} \quad \text{diag}(\mC) = 0
\end{aligned}
\end{equation}
where $\Vert \cdot \Vert_1$ denotes the element-wise $\ell_1$ norm (sum of absolute values) and the constraint $\text{diag}(\mC)=0$ prevents a point from representing itself.
\subsection{Scalable Subspace Clustering} 
In \cite{fettal2022subspace,fettal2023scalable,fettal2023boosting}, the authors proposed a scalable low-rank subspace clustering model. Specifically, the problem they considered is the following:
\begin{equation}
\begin{aligned}
\min_{\mU} \quad & \Vert \mX-\mU\mU^\top\mX\Vert^2 \quad \text{s.t.} \quad \mU^\top\mU=\mI.
\end{aligned}
\label{eq:problem-sagsc}
\end{equation}
To address the problem they use nonnegative kernel maps to create an affinity matrix
    $\mW=\Phi(\mU)\Phi(\mU)^\top.$
The advantage of computing the affinity matrix in this manner is that spectral partitioning can be obtained by singular values decomposition on $\Phi(\mU)$ instead of performing eigendecomposition on $\mW$, leading to substantial gains in spatial and computational efficiencies.

\subsection{Multi-view Subspace Clustering} 
Given a set of features $\mX_1, \ldots, \mX_V$ describing the same data points, a rudimentary multi-view subspace clustering model may consider a problem of the following form:
\begin{equation}
\begin{aligned}
    \min_{\mF,\mC_1,\ldots,\mC_V} \quad &
    \sum_v \Vert \mX_v - \mC_v\mX_v\Vert^2 
    +
    \texttt{Tr}( \mF^\top\mC_v\mF)\\
    \text{s.t.} \quad &
    \mF\in\{0,1\}^{n\times k}, \quad 
    \mF\mathbf{1}=\mathbf{1}, \quad 
    \mC_1,\ldots,\mC_V \in \mathcal{C},
\end{aligned}
\end{equation}
where $\mC_v$ is the coefficient matrix of the $v$-th view and $\mF$ is a consensus partition matrix. However, although the block structures in different view-specific coefficient matrices are similar, the magnitudes and signs of the entries in each $C_v$ can vary greatly. To solve this problem, \cite{gao2015multi} proposed using affinity matrices $\mW_v$ rather than coefficient matrices $\mC_v$. We generalize their formulation by eliminating the specific constraints that the authors used. This results in the following generic multi-view subspace clustering problem:
\begin{equation}
\begin{aligned}
    \min_{\mF,\mC_1,..,\mC_V} &\quad 
    \sum_v \Vert \mX_v - \mC_v\mX_v\Vert^2
     +
    \texttt{Tr}\left(\mF^\top\left(\mI-\tfrac{|\mC_v|+|\mC_v^\top|}{2}\right)\mF\right)\\
    \text{s.t.} \quad 
    &\mF\in\{0,1\}^{n\times k}, \quad 
    \mF\mathbf{1}=\mathbf{1}, \quad 
     \mC_1,\ldots,\mC_V \in \mathcal{C}.
    \label{eq:problem-multiview-original}
\end{aligned}
\end{equation}
Once solutions $\mC_1, \ldots, \mC_V$ are available, depending on the regularization used, spectral clustering can be performed on $\mW=\sum_v\mC_v$ to obtain the partition matrix $\mF$. 

\section{Proposed approach}
\textcolor{black}{Starting from the multi-view framework presented in problem (\ref{eq:problem-multiview-original}) and, as in (\ref{eq:problem-sagsc}), we introduce low-rank constraints, also using the same affinity matrix. However, here we consider a symmetrically normalized affinity matrix rather than the unnormalized one. The use of such a matrix can improve the stability, interpretability, performance, and convergence of clustering algorithms (see, for example, \cite{ng2001spectral}). We therefore propose the following objective function:}
\begin{equation}
\begin{aligned}
    \min_{\mF,\mU_1,\ldots,\mU_V} \quad &
    \sum_v \Vert \mX_v - \mU_v\mU_v^\top\mX_v\Vert^2  + \ \texttt{Tr}( \mF^\top(\mI-\mW_v)\mF)\\
    \text{such that} \qquad & \mW_v=\mD_v^{-\frac{1}{2}}\Phi(\mU_v)\Phi(\mU_v)^\top\mD_v^{-\frac{1}{2}}, \qquad
    \mF\in\{0,1\}^{n\times k}, \quad \mF\mathbf{1}=\mathbf{1}
    \label{eq:problem1}
\end{aligned}
\end{equation}
where \textcolor{black}{$\mU_v \in \mathbb{R}^{n\times f}$ is a low-rank matrix of view $v$ ($f << d$), and $\mW_v$ is the normalized affinity matrix of view $v$; the degree matrix $D_v$ is a diagonal matrix in which each diagonal element represents the degree of the corresponding node $i$ in the graph of view $v$. If we were to address this problem directly, the consensus partition $\mF$ would be derived through spectral clustering on}
%
\begin{equation}
\mW = \sum_v \mW_v = \sum_v \mD_v^{-\frac{1}{2}}\Phi(\mU_v)\Phi(\mU_v)^\top\mD_v^{-\frac{1}{2}}. \label{eq:consensus-affinity-matrix}    
\end{equation}
\textcolor{black}{However, applying spectral clustering through the proper decomposition of this matrix would lead to a computational complexity of}
$\mathcal{O}(n^2k)$ and a spatial complexity of $\mathcal{O}(n^2) $. This \textcolor{black}{implies} that we would lose the efficiency argued by \cite{fettal2023scalable} and thus lose the advantage of using such an approach in the first place. Therefore, we propose \textcolor{black}{addressing} this problem via the summation property of kernels.

\subsection{Multi-view Scalability via Kernel Summation}
Since we are using kernels to compute the view-specific affinity matrices $\mC_1,\ldots,\mC_V$, we propose using another property of kernels in order to obtain a factorized matrix consensus affinity matrix, which will then allow for efficient spectral clustering. Consider the following: 
\begin{proper}
Given two kernels $k_a$ and $k_b$ then their summation $k(x, y)=k_a(x, y)+k_b(x,y)$ is also a kernel.
\end{proper}
This means that matrix (\ref{eq:consensus-affinity-matrix}) is also a kernel and, consequently, has a decomposition of the form $\mB\mB^\top$ for some matrix $\mB$ (this is the property of kernel matrices). Additionally, we have the obvious fact:
\begin{proper}
When $k_a$ and $k_b$ are non-negative kernels, i.e. $\forall x,y$ $k_a(x, y)\geq 0$ and $k_b(x, y)\geq 0$, then $k(x, y) = k_a(x, y)+k_b(x,y)$ is also a nonnegative kernel.
\end{proper}
This means that every entry of (\ref{eq:consensus-affinity-matrix}) is non-negative, making it a valid affinity matrix. 
To solve the efficiency issue of combining the different view-specific affinity matrices, we therefore need a way of obtaining a vector realization $\mB$ for the decomposition of (\ref{eq:consensus-affinity-matrix}). Once we have $\mB$, we can obtain a spectral clustering by decomposition of singular values of $\mB$ instead of the eigendecomposition of $\mB\mB^\top$. To this end, we make use of the following property:
\begin{proper}
\label{prop:summation-kernel}
Given kernels $k_a$ and $k_b$ and their associated feature maps $\Phi_a$ and $\Phi_b$, the feature map corresponding to their summation kernel $k$ is $\Phi(x)=\Phi_a(x) \oplus \Phi_b(x)$.
\end{proper}
Using this property, we find that a realization is $\mB=\left[\mD_1^{-\frac{1}{2}}\Phi(\mU_1),\ldots, \mD_V^{-\frac{1}{2}}\Phi(\mU_V)\right]$. This decomposition means that the complexity of the approach is proportional to dimension $m$ of the feature map $\Phi$. This leads to large speedup and memory saving when $m << n$.

\subsection{Weighing Views with respect to their Clustering Structure} 
We additionally introduce a weighting scheme on the view-specific terms in problem (\ref{eq:problem1}), since different views should not always have the same contribution level towards the consensus affinity matrix. Here, we evaluate the importance of each view using the affinity matrix corresponding to that view. \textcolor{black}{To be more precise, the proposed objective function changes into, with the same constraints:}
%
\begin{equation}
\begin{aligned}
    \min_{\mF,\mU_1,\ldots,\mU_V} \quad &
    \sum_v \Vert \mX_v - \mU_v\mU_v^\top\mX_v\Vert^2\  
     + \ \lambda_v\ \ \texttt{Tr}( \mF^\top(\mI-\mW_v)\mF)
    \\\text{s.t.} \quad & \mW_v=\mD_v^{-\frac{1}{2}}\Phi(\mU_v)\Phi(\mU_v)^\top\mD_v^{-\frac{1}{2}},\quad\mF\in\{0,1\}^{n\times k}, \quad \mF\mathbf{1}=\mathbf{1}
\end{aligned}
\end{equation}
where we define the regularization vector $\boldsymbol\lambda=[\lambda_1,\ldots,\lambda_V]$ with respect to the cluster-ability of the $v$-th view: 
\begin{equation}
\begin{aligned}\label{eq:lambda}
    &\qquad\lambda_v = \frac{\texttt{exp}(\ \texttt{Tr}(\ \mG_v^\top(\mI-\mW_v)\mG_v)\ / \ T\ )}{\sum_w \texttt{exp}(\ \texttt{Tr}(\ \mG_w^\top(\mI-\mW_w)\mG_w)\ /\ T\ )}\\
     \text{s.t. } & \forall v\in [V]\ \mW_v=\Phi(\mU_v)\Phi(\mU_v)^\top, 
    \quad\mG_v\in\{0,1\}^{n\times k}, \mG_v\mathbf{1}=\mathbf{1} 
\end{aligned}
\end{equation}
where $T$ is a temperature parameter and $\mG_1,\ldots,\mG_V$ are different view-specific partition matrices. 

\textcolor{black}{We solve this problem via spectral clustering: we apply singular value decomposition on $\mD_v^{-\frac{1}{2}}\Phi(\mU_v)$ and then apply a clustering algorithm on the resulting left singular vectors, repeating the process for each $v \in [V]$.} 


\subsection{Optimization}
\textcolor{black}{In addition to the use of properties associated with kernel feature maps, the efficiency of our approach also stems from the fact that we do not use an iterative algorithm, in contrast to many other state-of-the-art models, e.g, \cite{pan2021multi}.} The algorithm consists of the following steps:
\subsubsection{Solving for $\mC_1,\ldots, \mC_V$} The first step is fixing\\ $\mC_1,\ldots, \mC_{v-1}, \mC_{v+1}, \ldots, \mC_V$ and $\mF$ in problem (\ref{eq:problem1}) and reiterating for each $v \in [V]$. The resulting problem is (\ref{eq:problem-sagsc}), and the solution is to set $\mU$ as the $f$ left singular vectors of $\mX$ associated with the largest singular values. Note that to solve our approach in one step and for scalability, we consider $\mW_v$ to be not a function of $\mC_v$.

\subsubsection{Computing $\lambda_1, \ldots, \lambda_V$} Once we have obtained the singular vectors associated with each coefficient matrix $\mC_v$, we compute the corresponding $\lambda_v$ using Formula (\ref{eq:lambda}).

\subsubsection{Solving for $\mF$} When fixing every $\mC_v$, we have to solve the following problem
\begin{equation}
\begin{aligned}
    \min_{\mF} \quad &
    \sum_v \texttt{Tr}( \mF^\top(\mI-\mW_v)\mF)
    \text{such that} \quad 
    \mF\in\{0,1\}^{n\times k}, \quad 
    \mF\mathbf{1}=\mathbf{1}.
\end{aligned}
\end{equation}
%
The solution, then, consists in using Property \ref{prop:summation-kernel}, and noticing that:
\begin{equation}
\begin{aligned}
\mW = & \sum_v \mW_v =  \sum_v \lambda_v\mD_v^{-\frac{1}{2}}\Phi(\mU_v)\Phi(\mU_v)^\top\mD_v^{-\frac{1}{2}}\\
= &\left(\oplus_v\ \sqrt{\lambda_v}\mD_v^{-\frac{1}{2}}\Phi(\mU_v)\right)\left(\oplus_v\ \sqrt{\lambda_v}\mD_v^{-\frac{1}{2}}\Phi(\mU_v)\right)^\top.
\end{aligned}
\end{equation}
The solution $\mF$ of this problem is obtained by applying $k$-means clustering on the left singular values corresponding to the largest $k$ singular values of matrix $\oplus_v\ \sqrt{\lambda_v}\mD_v^{-\frac{1}{2}}\Phi(\mU_v)$. A pseudo-code for our approach is given in Algorithm \ref{alg:pseudocode}.

\begin{algorithm}[!h]
\begin{algorithmic}[1]
\caption{\bf Pseudocode for MvSCK}
\label{alg:pseudocode}
\REQUIRE{
Data matrices $\mX_1, \ldots, \mX_V$, 
number of clusters $k$, 
number of components $f$, 
kernel feature map $\Phi$,
temperature $T$
}
\ENSURE{
Consensus partition matrix $\mF$
}
\FOR{ $v \in [V]$}
\STATE Subtract column means from $\mX_v$
\STATE Construct matrix $\mU_v$ by stacking the $f$ first left singular vectors of $\mX_v$ as its columns
\STATE $\mB_v = \Phi(\mU_v)$
\STATE Compute $d$ the row sums of $\mB_v\mB_v^\top$ efficiently
\STATE $\mB_v = \mB_v * \textBF{d}^{-0.5}$
\STATE Construct matrix $\mQ_v$ by stacking the $f+1$ first left singular vectors of $\mB_v$ as its columns and then drop the first column
\STATE Compute partition matrix $\mG_v$ via k-means on $\mQ_v$ with $k$ clusters
\ENDFOR
\STATE Compute $\lambda_1,...,\lambda_V$ using Formula (\ref{eq:lambda}) with temperature $T$, $ \mG_v$ and $\mW_v=\mB_v\mB_v^\top$
\STATE $\mB = \oplus_v \lambda_v\mB_v $
\STATE Compute $d$ the row sums of $\mB\mB^\top$ efficiently
\STATE $\mB = \mB * \textBF{d}^{-0.5}$
\STATE Construct matrix $\mQ\phantom{v}$ by stacking the $f+1$ first left singular vectors of $\mB\phantom{v}$ as its columns and then drop the first column
\STATE Compute partition matrix $\mF$ via k-means on $\mQ$ with $k$ clusters
\end{algorithmic}
\end{algorithm}

\subsection{Complexity Analysis} 
To complement our intuitive justification for the efficiency of our approach, we give a theoretical analysis of its complexity:
\begin{enumerate}[leftmargin=.66cm]
\item \textBF{Solving for $\mC_1,\ldots, \mC_V$} This process takes $O(vnd\log(f))$ where $f$ is the chosen rank due to the randomized singular values decomposition.
\item \textBF{Computing $\lambda_1, \ldots, \lambda_V$} The complexity corresponds to computing the projections using feature map $\Phi$. We consider the case of using a quadratic polynomial feature map whose computational complexity is $O(nf^2)$. The computational complexity of this step is then $O(Vnf^2log(k))$.
\item \textBF{Solving for $\mF$} Finally, the process of computing the concatenated matrix $\mB$ and then applying singular value decomposition is $O(Vnf^2\log(k))$. Applying k-means afterwards results in $O(tVnf^2k)$ operations  \textcolor{black}{where $t$ denotes the number of iterations}.
\item \textBF{The effect of the choice of the non-negative Kernel} It is possible to use nonnegative kernels other than the quadratic polynomial. However, in the case of infinite-dimensional kernels such as RBF and sigmoid kernels or, in the case of polynomials, kernels with large even degrees, it is necessary to use kernel feature map approximation techniques such as Nystroem \cite{williams2000using} or polynomial sketching \cite{pham2013fast}. In these cases, the factor $f^2$ is replaced by the dimension of the feature map approximation, and the complexity of computing the approximation of the kernel has to be added to the total complexity of our approach.
\end{enumerate}
In table \ref{tab:complexity} we report the complexity of several state-of-the-art algorithms. We see that considerably more factors must be taken into account than are needed when using our approach. Note that we have added the complexity of learning graph representations (using (\ref{eq:laplacian-smoothing})) to non-graph methods.

\begin{table}[h]
\caption{Complexity of some of the considered approaches on network data. $m$ represents the number of anchors. $|E|$ is the number of edges in the graphs used in Formula (\ref{eq:laplacian-smoothing}). For the sake of simplicity, we consider that the dimension of the features and the number of graph edges in the different views are the same.}
\label{tab:complexity}
\centering
\begin{tabular}{@{}lcc@{}}
\toprule
\textBF{Method} & \textBF{Time Complexity} \\\midrule
LMVSC& $\mathcal{O}(Vp|E|d+nm^3V+m^3V^3 + mVn + nk^2t)$ \\
FDAGF& $\mathcal{O}(Vp|E|d+rVdm^2 + rm^2n + rVdmn)$\\
SMVSC& $\mathcal{O}(Vp|E|d+Vd^3 + Vd^2k^2 +md^2 + dmk^2 + nm^3)$ \\
MAGC & $\mathcal{O}(Vn^2d)$\\
MvAGC & $\mathcal{O}(tm^3V+tVm^2nd)$\\
LMGEC& $\mathcal{O}(V|E|d + Vndk)$\\\midrule
\textBF{MvSCK}&  $O(Vp|E|d + Vnf^2\log(k))$\\
\bottomrule
\end{tabular}
\end{table}


\section{Experiments}
In the following comparisons, our approach is identified with the label \textBF{MvSCK} (for \textit{\textBF{M}ulti-\textBF{v}iew \textBF{S}ubspace \textBF{C}lustering with explicit \textBF{K}ernel feature maps}). We aim to answer the following research questions:
\begin{itemize}[leftmargin=.35cm]
\item How does MvSCK compare with state-of-the-art in terms of: 
\begin{itemize}[leftmargin=.5cm]
    \item \textBF{RQ1}: Clustering performance? 
    \item \textBF{RQ2}: Computational efficiency?
\end{itemize} 
\item \textBF{RQ3:} Does our approach consistently outperform the state-of-the-art ?
\item \textBF{RQ4:} How do the hyper-paramters affect MvSCK? 
\item \textBF{RQ5:} Is the view-importance parameter necessary?
\end{itemize}

Possible future research questions which could be explored in the future include extension to co-clustering \cite{fettal2022subspace,fettal2023boosting}, incomplete-views \cite{xu2015multi,jiang2024deep} and streaming data \cite{shamsinejad2025providing,shamsinejad2024anonymizing}, etc.
\subsection{Datasets}

We evaluate our method on six benchmark datasets with different degrees of overlap: two small scale datasets, namely ACM and DBLP \cite{fan2020one2multi}; two medium datasets, namely Photos and Computers \cite{shchur2018pitfalls}; and two large datasets ArXivand Products \cite{hu2020open}. \textcolor{black}{The characteristics of the datasets are reported in Table \ref{tab:datasets}.}  
The six datasets are described briefly below:
\begin{itemize}[leftmargin=.35cm]
\item\textBF{ACM} is a network dataset of research papers, where the associated node features are the papers' keywords in bag-of-words format. The ``Co-Subject'' and ``Co-Author'' graphs respectively connect papers with the same theme or the same author. 
\item\textBF{DBLP} is a network dataset of authors with bag-of-words node features representing keywords. ``Co-Author'', ``Co-Conference'', and ``Co-Term'' links indicate that the authors concerned have collaborated on papers, published papers at the same conference, or published papers with the same terms.
\item\textBF{Photos \& Computers} are segments of the Amazon co-purchase network dataset. The nodes represent goods, and their attribute features are represented by bag-of-words of product reviews. The edges indicate that two goods are purchased together. The second set of attributes represents the word co-occurrence matrix. 
\item\textBF{OGBN-ArXiv} represents a citation network encompassing a set of computer science arXiv papers. Each paper comes with a 128-dimensional feature vector obtained by averaging the embeddings of words in its title and abstract. We add a second $k$-nearest neighbors (k-NN) graph with 10 neighbors and added self-connections.
\item\textBF{Products} is a graph representing an Amazon product co-purchase network. An edge between two products signifies that those products are purchased together. The node features are created by extracting bag-of-words features from the product descriptions and reducing their dimensionality to 100 via PCA. Similarly, to OGBN-ArXiv, we add a second k-NN graph.
\end{itemize}
The statistical summaries for the different datasets are given in Table \ref{tab:datasets}. 

\begin{table}[h!]
\caption{Characteristics of the Datasets. \textcolor{black}{For example, ArXivrepresents a citation network consisting of a set of computer science arXiv papers. Each paper is associated with a 128-dimensional feature vector obtained by averaging the embeddings of words in its title and abstract. We also add a second kk-nearest neighbors (k-NN) graph with 10 neighbors and include self-connections.} The imbalance is the ratio between the most frequent and least frequent classes.}
\label{tab:datasets}
\centering
\scriptsize{
\begin{tabular}{@{}lccccccc@{}}
\toprule
Dataset & Scale & Nodes & Features & Graph and Edges & Clusters & Imbalance \\ \midrule
ACM   & Small & 3,025 & 1,830 & \begin{tabular}{c}
     Co-Subject (29,281)\\\hline
     Co-Author (2,210,761)
\end{tabular} & 3 & 1.1\\\hline
DBLP & Small  &4,057 & 334 & \begin{tabular}{c}
     Co-Author (11,113)\\\hline
     Co-Conference (5,000,495)\\\hline
     Co-Term (6,776,335)
\end{tabular} & 4 & 1.6\\\hline
Photos & Medium &7,487 & \begin{tabular}{c}
     745\\\hline
     7,487
\end{tabular} & Co-Purchase (119,043) & 8 & 5.7\\ \bottomrule
Computers & Medium & 13,381 & \begin{tabular}{c}
     767\\\hline
     13,381
\end{tabular} & Co-Purchase (504,937) & 10 & 17.5\\ \bottomrule
ArXiv& Large &169,343 & 128 & \begin{tabular}{c}
     k-NN (1,862,773)\\\hline
     Co-Citation (2,484,941)\\
\end{tabular} & 40 & 942.1\\\hline
Products & Large &2,449,029 & 100 & \begin{tabular}{c}
     k-NN (26,939,319)\\\hline
     Co-Purchase (126,167,053)\\
\end{tabular} & 47 & 668,950\\\hline
\end{tabular}
}
\end{table}

\subsection{Experimental Setup}

\subsubsection{ Experimental setting} For the purposes of comparison, we validate our approach against the state-of-the-art approaches discussed in the Related Work section above, with reference to clustering accuracy (CA), clustering F1-score (CF1) \cite{powers2020evaluation}, adjusted rand index (ARI) \cite{steinley2004properties}, and normalized mutual information (NMI) \cite{strehl2002cluster}. We use the official implementations of the compared approaches. We choose parameters according to the authors' recommendations. When these recommendations are not available, we set parameters similarly to how they were set on datasets of similar dimensions. To make comparisons fair, we remove any supervised signals when running experiments in the original code; these would include, for example, the validation of runs based on supervised metrics such as accuracy or setting random seeds. 
\textcolor{black}{In our method, we employ the quadratic kernel, which was also utilized in \cite{fettal2023scalable} and demonstrated to be effective by the authors.} The number of components parameter $f$ was set equal to the number of clusters in all datasets, with the exception of Products, where it was set to 10 due to its large size. Note that all metrics reported are averaged over five different runs.
All experiments were performed on a laptop equipped with a 12th Gen Intel(R) Core(TM)
i9-12950HX (24 CPU threads) 2.3 GHz processor, a 12GB GeForce RTX 3080 Ti GPU, and 64GB of RAM.

\subsubsection{ Learning Node Representations}
To perform clustering on network datasets using generic approaches such as ours in addition to LMVSC, SMVSC, FDAGF and OMSC, we first need to learn node embeddings. Given a multi-view attributed network consisting of adjacency matrices $\mA_1,\ldots,\mA_V$ and feature matrices $\mX_1,\ldots,\mX_V$, we learn embeddings via Laplacian smoothing. This is a well-known technique that has been used in works such as \cite{pan2021multi,fettal2023simultaneous} to handle graph data. It involves consecutive neighborhood averaging steps:
\begin{equation}
    \forall v \in [V] \qquad \mX_v \gets \mA_v^p\mX_v
    \label{eq:laplacian-smoothing}
\end{equation}
where $p$ is the propagation order. Specifically, in our experiments, we set the propagation order as follows: 2 for ACM, 2 for DBLP, 20 for Photos, 40 for Computers, 60 for OGBN-ArXiv, and 100 for Products.  This step can be seen as a preprocessing step that is independent of the approaches themselves. It is a methodology that has emerged from work done on simplifying graph convolutional networks \cite{wu2019simplifying}.

\subsection{Clustering Results (RQ1)}

The clustering outcomes for small, medium, and large graph datasets are detailed in Tables \ref{tab:small}, \ref{tab:medium}, and \ref{tab:large} respectively. A comprehensive analysis of these tables reveals the superior performance of our approach relative to other methodologies across a majority of datasets and evaluation metrics. Remarkably, our method exhibits superiority even when compared to specialized techniques designed explicitly for graph datasets. Noteworthy observations include LMVSC's competitiveness with our approach on the Computers dataset, where both methods achieve comparable results. Basically, while LMVSC outperforms our approach in terms of F1 and ARI metrics, our method outperforms it in accuracy and NMI metrics. This comparison underscores the strengths of our approach, showcasing its robust performance across diverse datasets and metrics. We observed some disparities between the results reported in the original papers and those obtained by running the code on the same datasets. Additionally, many approaches struggle to scale to challenging datasets like ArXivand Products. Specifically, OMSC fails to retrieve the correct number of clusters in Photos and Computers, impacting its ability to compute CF1 scores.

\begin{table}[h]
\caption{Clustering results on the small scale datasets.}
\label{tab:small}
\centering
\scriptsize{
\begin{tabular}{@{}llllllllll@{}}
\toprule
{} & \multicolumn{4}{c}{\textBF{ACM}} & \multicolumn{4}{c}{\textBF{DBLP}} \\\cmidrule(lr){2-5}\cmidrule(lr){6-9}
{} & CA & CF1 & NMI & ARI & CA & CF1 & NMI & ARI  \\
\midrule
LMVSC        &           90.33\tiny{ ±0.0} &           90.59\tiny{ ±0.1} &           68.33\tiny{ ±0.2} &           73.64\tiny{ ±0.1} &           52.72\tiny{ ±0.0} &           58.30\tiny{ ±0.0} &           20.15\tiny{ ±0.0} &           14.97\tiny{ ±0.0} \\
OMSC & 70.21\tiny{ ±0.0} &           71.11\tiny{ ±0.0} &            45.81\tiny{ ±0.0} &            43.55\tiny{ ±0.0} & 
91.45\tiny{ ±0.0} &           90.95\tiny{ ±0.0} &           74.59\tiny{ ±0.0} &           79.82\tiny{ ±0.0} \\
FDAGF        &           70.27\tiny{ ±0.0} &           79.39\tiny{ ±0.0} &           47.07\tiny{ ±0.1} &           45.20\tiny{ ±0.1} &           89.61\tiny{ ±0.0} &           89.39\tiny{ ±0.0} &           71.56\tiny{ ±0.0} &           75.78\tiny{ ±0.0} \\
SMVSC        &           70.45\tiny{ ±0.0} &           71.80\tiny{ ±0.0} &           46.35\tiny{ ±0.0} &           43.91\tiny{ ±0.0} &           91.45\tiny{ ±0.0} &           90.87\tiny{ ±0.0} &           74.50\tiny{ ±0.0} &           79.88\tiny{ ±0.0} \\
MvAGC        &           83.35\tiny{ ±4.9} &           83.49\tiny{ ±5.0} &           55.86\tiny{ ±7.1} &           58.78\tiny{ ±8.6} &           76.62\tiny{ ±6.4} &           67.10\tiny{ ±3.5} &          59.99\tiny{ ±10.5} &           59.87\tiny{ ±6.5} \\
MAGC         &           55.81\tiny{ ±7.8} &           47.58\tiny{ ±4.1} &          26.09\tiny{ ±12.6} &          20.05\tiny{ ±16.6} &           91.93\tiny{ ±0.6} &           91.43\tiny{ ±0.6} &           75.28\tiny{ ±1.1} &           80.53\tiny{ ±1.4} \\
MCGC         &           89.92\tiny{ ±0.0} &           89.93\tiny{ ±0.0} &           67.03\tiny{ ±0.0} &           72.56\tiny{ ±0.0} &           87.63\tiny{ ±0.0} &           77.24\tiny{ ±9.9} &           69.37\tiny{ ±2.0} &           79.24\tiny{ ±7.9} \\
LMGEC        &           93.00\tiny{ ±0.0} &           93.06\tiny{ ±0.0} &  \textBF{75.13\tiny{ ±0.1}} &           80.27\tiny{ ±0.1} &           92.85\tiny{ ±0.0} &           92.37\tiny{ ±0.0} &           77.40\tiny{ ±0.0} &           82.84\tiny{ ±0.0} \\
\midrule
\textBF{MvSCK} &  \textBF{93.21\tiny{ ±0.1}} &  \textBF{93.22\tiny{ ±0.1}} &           74.97\tiny{ ±0.1} &  \textBF{80.76\tiny{ ±0.1}} &  \textBF{93.09\tiny{ ±0.1}} &  \textBF{92.57\tiny{ ±0.1}} &  \textBF{78.05\tiny{ ±0.2}} &  \textBF{83.36\tiny{ ±0.2}} \\
\bottomrule
\end{tabular}
}
\end{table}

\begin{table}[h]
\caption{Clustering results on the medium scale datasets.}
\label{tab:medium}
\centering
\scriptsize{
\begin{tabular}{@{}llllllllll@{}}
\toprule
{} & \multicolumn{4}{c}{\textBF{Photos}} & \multicolumn{4}{c}{\textBF{Computers}} \\\cmidrule(lr){2-5}\cmidrule(lr){6-9}
 & CA & CF1 & NMI & ARI & CA & CF1 & NMI & ARI   \\
\midrule
LMVSC        &           61.23\tiny{ ±1.2} &           55.87\tiny{ ±1.4} &           57.17\tiny{ ±0.7} &           36.60\tiny{ ±0.7} &           64.97\tiny{ ±0.9} &  \textBF{62.24\tiny{ ±3.4}} &           52.39\tiny{ ±0.9} &  \textBF{48.76\tiny{ ±0.1}} \\
OMSC & 61.44\tiny{ ±0.0} &  N/A &            49.28\tiny{ ±0.0} &           36.73\tiny{ ±0.0} &
63.99\tiny{ ±0.0} &  N/A &           42.27\tiny{ ±0.0} &           42.89\tiny{ ±0.0} \\
FDAGF        &           55.48\tiny{ ±0.1} &           49.37\tiny{ ±0.0} &           52.96\tiny{ ±0.1} &           30.01\tiny{ ±0.1} &           64.68\tiny{ ±4.1} &           56.93\tiny{ ±0.8} &           52.82\tiny{ ±0.4} &           44.94\tiny{ ±2.6} \\
SMVSC        &           61.67\tiny{ ±0.1} &           55.74\tiny{ ±0.3} &           50.83\tiny{ ±0.2} &           35.99\tiny{ ±0.1} &           64.72\tiny{ ±0.1} &           49.55\tiny{ ±0.6} &           49.09\tiny{ ±0.4} &           46.01\tiny{ ±0.1} \\
MvAGC        &           45.28\tiny{ ±6.7} &           43.15\tiny{ ±4.4} &           29.96\tiny{ ±5.9} &           20.02\tiny{ ±3.9} &           45.82\tiny{ ±4.1} &           45.12\tiny{ ±4.8} &           28.45\tiny{ ±5.5} &           18.98\tiny{ ±3.1} \\
MAGC         &           52.36\tiny{ ±2.7} &           43.76\tiny{ ±3.5} &           48.86\tiny{ ±2.5} &           21.42\tiny{ ±2.3} &           63.48\tiny{ ±6.1} &          51.81\tiny{ ±10.2} &          52.47\tiny{ ±10.1} &           35.48\tiny{ ±9.4} \\
MCGC         &           41.86\tiny{ ±3.5} &           26.75\tiny{ ±7.7} &          15.16\tiny{ ±10.6} &           23.61\tiny{ ±9.4} &           45.29\tiny{ ±0.7} &           34.17\tiny{ ±8.2} &           22.24\tiny{ ±7.4} &          26.96\tiny{ ±10.4} \\
LMGEC        &           70.93\tiny{ ±0.0} &           64.83\tiny{ ±0.0} &           60.88\tiny{ ±0.0} &           51.05\tiny{ ±0.0} &           46.10\tiny{ ±1.5} &           40.21\tiny{ ±0.8} &           46.14\tiny{ ±0.9} &           27.61\tiny{ ±0.4} \\
\midrule    
\textBF{MvSCK} &  \textBF{75.66\tiny{ ±2.2}} &  \textBF{66.56\tiny{ ±5.0}} &  \textBF{70.66\tiny{ ±1.3}} &  \textBF{58.65\tiny{ ±2.0}} &  \textBF{66.34\tiny{ ±1.5}} &           56.23\tiny{ ±0.8} &  \textBF{57.15\tiny{ ±0.8}} &           46.27\tiny{ ±2.0} \\
\bottomrule
\end{tabular}
}
\end{table}

\begin{table}[h!]
\caption{Clustering results on the large scale datasets. We allow for a maximum runtime of two hours per run.}
\label{tab:large}
\centering
\scriptsize{
\begin{tabular}{@{}llllllllll@{}}
\toprule
{} & \multicolumn{4}{c}{\textBF{OGBN-ArXiv}} & \multicolumn{4}{c}{\textBF{Products}} \\\cmidrule(lr){2-5}\cmidrule(lr){6-9}
 & CA & CF1 & NMI & ARI & CA & CF1 & NMI & ARI   \\
\midrule
LMVSC        &           30.94\tiny{ ±0.6}&      23.51\tiny{ ±4.6}&      31.92\tiny{ ±1.0}&      20.07\tiny{ ±0.5} & \multicolumn{4}{c}{Timeout} \\
OMSC        &           \multicolumn{4}{c}{Timeout} & \multicolumn{4}{c}{Timeout} \\
FDAGF        &           \multicolumn{4}{c}{Timeout} & \multicolumn{4}{c}{Timeout} \\
SMVSC        &           \multicolumn{4}{c}{Timeout} & \multicolumn{4}{c}{Timeout} \\
MvAGC        &            \multicolumn{4}{c}{OOM} & \multicolumn{4}{c}{OOM} \\
MAGC         &           \multicolumn{4}{c}{OOM} & \multicolumn{4}{c}{OOM} \\
MCGC         &           \multicolumn{4}{c}{Timeout} & \multicolumn{4}{c}{Timeout} \\
LMGEC&29.61\tiny{ ±0.8}&16.44\tiny{ ±0.7}&34.53\tiny{ ±0.3}&19.86\tiny{ ±0.7}&36.09\tiny{ ±1.1}&16.23\tiny{ ±0.7}&43.64\tiny{ ±0.7}&21.60\tiny{ ±0.5}\\\midrule
\textBF{MvSCK}&\textBF{48.29\tiny{ ±1.8}}&\textBF{28.46\tiny{ ±0.9}}&\textBF{46.99\tiny{ ±0.5}}&\textBF{39.15\tiny{ ±1.2}}&\textBF{47.14\tiny{ ±1.5}}&\textBF{16.89\tiny{ ±1.3}}&\textBF{50.80\tiny{ ±0.8}}&\textBF{32.24\tiny{ ±1.3}}\\
\bottomrule
\end{tabular}
}
\end{table}

\subsection{Running Times (RQ2)}
The detailed breakdown of running times for different models across all datasets is presented in Table \ref{tab:times}, where our approach emerges as the leader with markedly shorter execution times over small, medium, and large graph datasets. The notable efficiency of our method becomes especially evident when compared to alternative approaches. Notably, for the challenging ArXivand Products datasets, our approach not only maintains its speed advantage but also adheres to the strict three-hour time limit we set. In contrast, several competing models exceed this limit, highlighting the computational efficiency of our proposed method in the face of diverse datasets.

\begin{table}[h]
\caption{Execution times on the different datasets (in seconds).}
\label{tab:times}
\centering
\scriptsize{
\begin{tabular}{@{}lllllll@{}}
\toprule
{} & \textBF{ACM} & \textBF{DBLP} &\textBF{Photos} & \textBF{Computers} &\textBF{ArXiv} & \textBF{Products} \\
\midrule
LMVSC        &           2.36\tiny{ ±0.1} & 2.97\tiny{ ±0.1}&           37.78\tiny{ ±3.7} &        144.26\tiny{ ±14.2} &2952.9\tiny{ ±210.4}&Timeout\\
OMSC         &          24.66\tiny{ ±0.2} &          27.51\tiny{ ±0.1} &        288.32\tiny{ ±43.1} &         414.94\tiny{ ±3.7} &Timeout&Timeout\\
FDAGF        &          37.31\tiny{ ±1.3} &          68.28\tiny{ ±3.2} &         382.76\tiny{ ±4.2} &       1555.28\tiny{ ±50.6} &Timeout&Timeout\\
SMVSC        &          16.94\tiny{ ±1.6} &          29.68\tiny{ ±0.6} &        169.55\tiny{ ±14.5} &        352.40\tiny{ ±95.5} &Timeout&Timeout\\
MvAGC        &           5.05\tiny{ ±0.1} &           5.65\tiny{ ±0.0} &          36.59\tiny{ ±1.0} &          36.76\tiny{ ±0.7} & OOM & OOM\\
MAGC         &          21.83\tiny{ ±1.6} &          38.12\tiny{ ±3.8} &        304.23\tiny{ ±22.6} &    8699.82\tiny{ ±16343.2} & OOM& OOM\\
MCGC         &         738.54\tiny{ ±4.5} &        358.56\tiny{ ±13.5} &      3803.44\tiny{ ±434.1} &    11393.04\tiny{ ±1690.1} &Timeout&Timeout\\
LMGEC        &           0.63\tiny{ ±0.4} &           0.51\tiny{ ±0.5} &           2.95\tiny{ ±0.3} &           7.80\tiny{ ±1.0} & 44.87\tiny{ ±1.4}
& 601.32\tiny{ ±29.2}\\\midrule
\textBF{MvSCK} &  \textBF{0.18\tiny{ ±0.1}} &  \textBF{0.19\tiny{ ±0.1}} &  \textBF{0.86\tiny{ ±0.1}} &  \textBF{5.96\tiny{ ±0.2}} & \textBF{20.43\tiny{ ±3.8}} & \textBF{239.32\tiny{ ±14.2}}\\
\bottomrule
\end{tabular}
}
\end{table}

\subsection{Statistical Significance Testing (RQ3)}
Figure \ref{fig:test-clustering} shows the results of the Holm mean-rank post-hoc test \cite{holm1979simple} over the clustering results. We compute the mean rank over rankings generated based on each (dataset, metric, run) triplet. We see that our approach is ranked first by a large margin, followed by LMGEC and LMVSC as second and third. After these three, there are various overlapping groups of approaches.

In Figure \ref{fig:test-times}, we report the results of the Holm mean-rank post hoc test on the execution times of the different algorithms. Ours is the most efficient, followed by LMGEC. LMVSC and MvAGC are ranked third, followed by SMVSC, OMSC, and then a group composed of MAGC and FDAGF. MCGC comes last. 

\begin{figure}[t]
\centering
\scriptsize{
\begin{tikzpicture}[
  treatment line/.style={rounded corners=1.5pt, line cap=round, shorten >=1pt},
  treatment label/.style={font=},
  group line/.style={ultra thick},
]
\begin{axis}[
  clip={false},
  axis x line={center},
  axis y line={none},
  axis line style={-},
  xmin={1},
  ymax={0},
  scale only axis={true},
  width={\axisdefaultwidth},
  ticklabel style={anchor=south, yshift=1.3*\pgfkeysvalueof{/pgfplots/major tick length}, font=},
  every tick/.style={draw=black},
  major tick style={yshift=.5*\pgfkeysvalueof{/pgfplots/major tick length}},
  minor tick style={yshift=.5*\pgfkeysvalueof{/pgfplots/minor tick length}},
  title style={yshift=\baselineskip},
  xmax={9},
  ymin={-5.5},
  height={6\baselineskip},
  xtick={1,3,5,7,9},
  minor x tick num={1},
]
\draw[treatment line] ([yshift=-2pt] axis cs:1.1666666666666667, 0) |- (axis cs:0.41666666666666674, -2.5)
  node[treatment label, anchor=east] {\textBF{MvSCK}};
\draw[treatment line] ([yshift=-2pt] axis cs:2.875, 0) |- (axis cs:0.41666666666666674, -3.5)
  node[treatment label, anchor=east] {LMGEC};
\draw[treatment line] ([yshift=-2pt] axis cs:4.333333333333333, 0) |- (axis cs:0.41666666666666674, -4.5)
  node[treatment label, anchor=east] {LMVSC};w
\draw[treatment line] ([yshift=-2pt] axis cs:5.354166666666667, 0) |- (axis cs:0.41666666666666674, -5.5)
  node[treatment label, anchor=east] {SMVSC};
\draw[treatment line] ([yshift=-2pt] axis cs:5.541666666666667, 0) |- (axis cs:7.625, -6.0)
  node[treatment label, anchor=west] {FDAGF};
\draw[treatment line] ([yshift=-2pt] axis cs:6.041666666666667, 0) |- (axis cs:7.625, -5.0)
  node[treatment label, anchor=west] {MAGC};
\draw[treatment line] ([yshift=-2pt] axis cs:6.145833333333333, 0) |- (axis cs:7.625, -4.0)
  node[treatment label, anchor=west] {OMSC};
\draw[treatment line] ([yshift=-2pt] axis cs:6.666666666666667, 0) |- (axis cs:7.625, -3.0)
  node[treatment label, anchor=west] {MCGC};
\draw[treatment line] ([yshift=-2pt] axis cs:6.875, 0) |- (axis cs:7.625, -2.0)
  node[treatment label, anchor=west] {MvAGC};
\draw[group line] (axis cs:5.354166666666667, -3.6666666666666665) -- (axis cs:5.541666666666667, -3.6666666666666665);
\draw[group line] (axis cs:5.541666666666667, -2.6666666666666665) -- (axis cs:6.145833333333333, -2.6666666666666665);
\draw[group line] (axis cs:6.041666666666667, -1.3333333333333333) -- (axis cs:6.875, -1.3333333333333333);
\end{axis}
\end{tikzpicture}
}
\caption{Holm post-hoc mean rank test ($\alpha=0.01$) with respect to clustering performance.}
\label{fig:test-clustering}
\end{figure}

\begin{figure}[h]
\centering
\scriptsize{
\begin{tikzpicture}[
  treatment line/.style={rounded corners=1.5pt, line cap=round, shorten >=1pt},
  treatment label/.style={font=},
  group line/.style={ultra thick},
]
\begin{axis}[
  clip={false},
  axis x line={center},
  axis y line={none},
  axis line style={-},
  xmin={1},
  ymax={0},
  scale only axis={true},
  width={\axisdefaultwidth},
  ticklabel style={anchor=south, yshift=1.3*\pgfkeysvalueof{/pgfplots/major tick length}, font=},
  every tick/.style={draw=black},
  major tick style={yshift=.5*\pgfkeysvalueof{/pgfplots/major tick length}},
  minor tick style={yshift=.5*\pgfkeysvalueof{/pgfplots/minor tick length}},
  title style={yshift=\baselineskip},
  xmax={9},
  ymin={-5.5},
  height={6\baselineskip},
  xtick={1,3,5,7,9},
  minor x tick num={1},
]
\draw[treatment line] ([yshift=-2pt] axis cs:1.0, 0) |- (axis cs:0.25, -2.5)
  node[treatment label, anchor=east] {\textBF{MvSCK}};
\draw[treatment line] ([yshift=-2pt] axis cs:2.0, 0) |- (axis cs:0.25, -3.5)
  node[treatment label, anchor=east] {LMGEC};
\draw[treatment line] ([yshift=-2pt] axis cs:3.8333333333333335, 0) |- (axis cs:0.25, -4.5)
  node[treatment label, anchor=east] {LMVSC};
\draw[treatment line] ([yshift=-2pt] axis cs:4.416666666666667, 0) |- (axis cs:0.25, -5.5)
  node[treatment label, anchor=east] {MvAGC};
\draw[treatment line] ([yshift=-2pt] axis cs:5.583333333333333, 0) |- (axis cs:8.833333333333334, -6.0)
  node[treatment label, anchor=west] {SMVSC};
\draw[treatment line] ([yshift=-2pt] axis cs:6.083333333333333, 0) |- (axis cs:8.833333333333334, -5.0)
  node[treatment label, anchor=west] {OMSC};
\draw[treatment line] ([yshift=-2pt] axis cs:6.75, 0) |- (axis cs:8.833333333333334, -4.0)
  node[treatment label, anchor=west] {MAGC};
\draw[treatment line] ([yshift=-2pt] axis cs:7.25, 0) |- (axis cs:8.833333333333334, -3.0)
  node[treatment label, anchor=west] {FDAGF};
\draw[treatment line] ([yshift=-2pt] axis cs:8.083333333333334, 0) |- (axis cs:8.833333333333334, -2.0)
  node[treatment label, anchor=west] {MCGC};
\draw[group line] (axis cs:3.8333333333333335, -3.0) -- (axis cs:4.416666666666667, -3.0);
\draw[group line] (axis cs:6.75, -2.0) -- (axis cs:7.25, -2.0);
\end{axis}
\end{tikzpicture}
}
\caption{Holm post-hoc mean rank test ($\alpha=0.01$) with respect to running times.}
\label{fig:test-times}
\end{figure}

\subsection{Sensitivity Analysis (RQ4)} 

\begin{figure}[h]
    \begin{subfigure}{\textwidth}
    \centering
    \tikz\node[inner sep=0pt,
           label=west:\rotatebox{90}{\textBF{ACM}}]{\hspace{.2cm}
      \centering
      \begin{subfigure}{.24\textwidth}
      \centering
      \includegraphics[width=\textwidth]{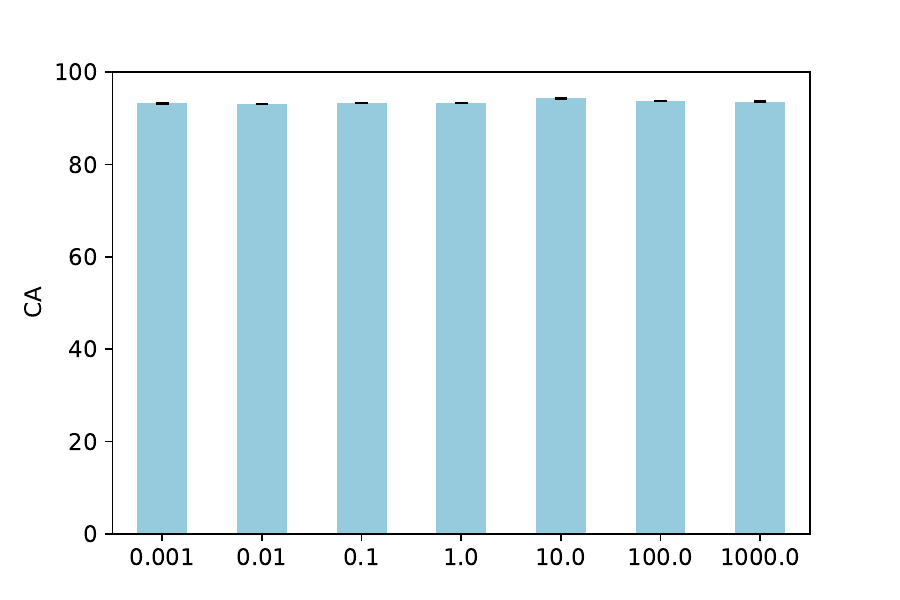}
    \end{subfigure}
    \begin{subfigure}{.24\textwidth}
      \centering
      \includegraphics[width=\textwidth]{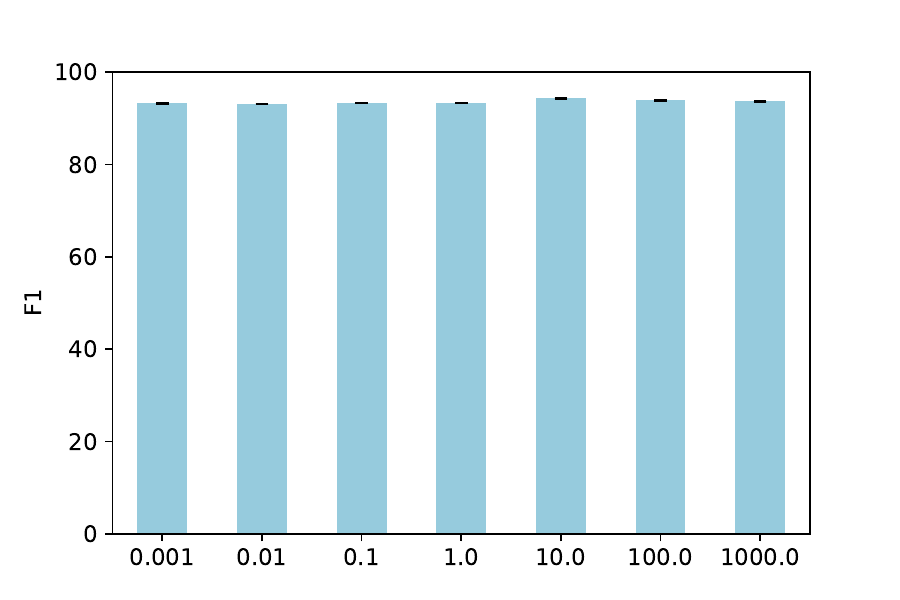}
    \end{subfigure}
    \begin{subfigure}{.24\textwidth}
      \centering
      \includegraphics[width=\textwidth]{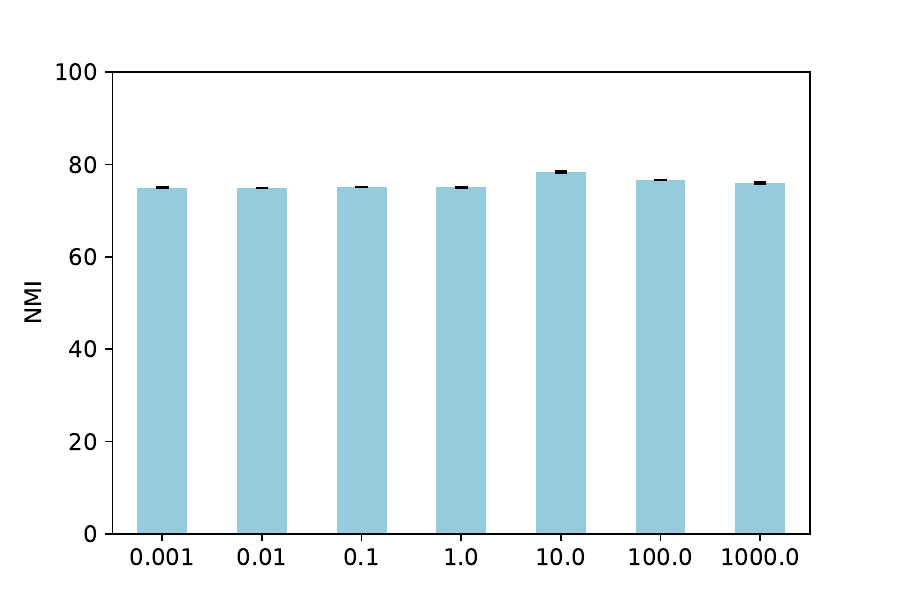}
    \end{subfigure}
    \begin{subfigure}{.24\textwidth}
      \centering
      \includegraphics[width=\textwidth]{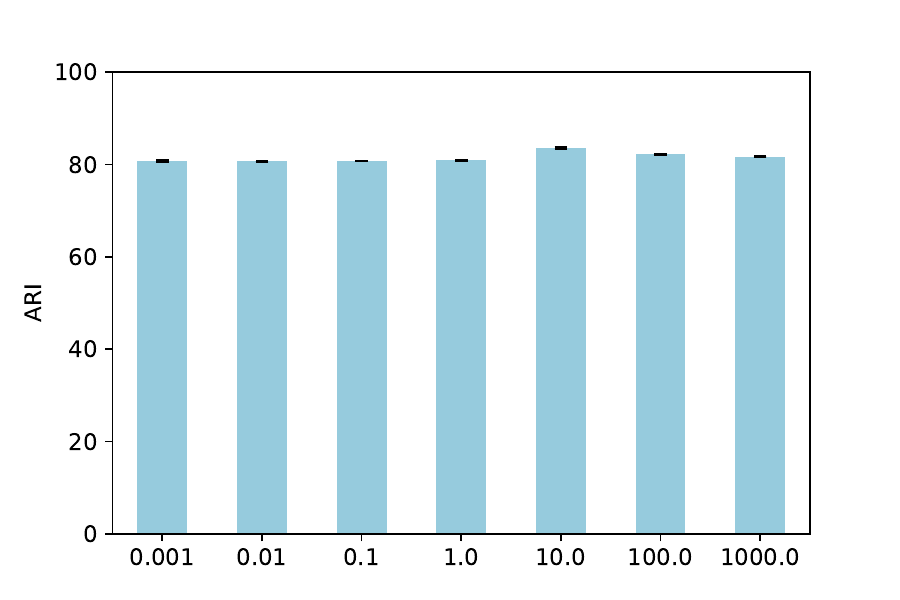}
    \end{subfigure}
    };
    \end{subfigure}
    \begin{subfigure}{\textwidth}
    \centering
    \tikz\node[inner sep=0pt,
           label=west:\rotatebox{90}{\textBF{DBLP}}]{\hspace{.2cm}
      \centering
      \begin{subfigure}{.24\textwidth}
      \centering
      \includegraphics[width=\textwidth]{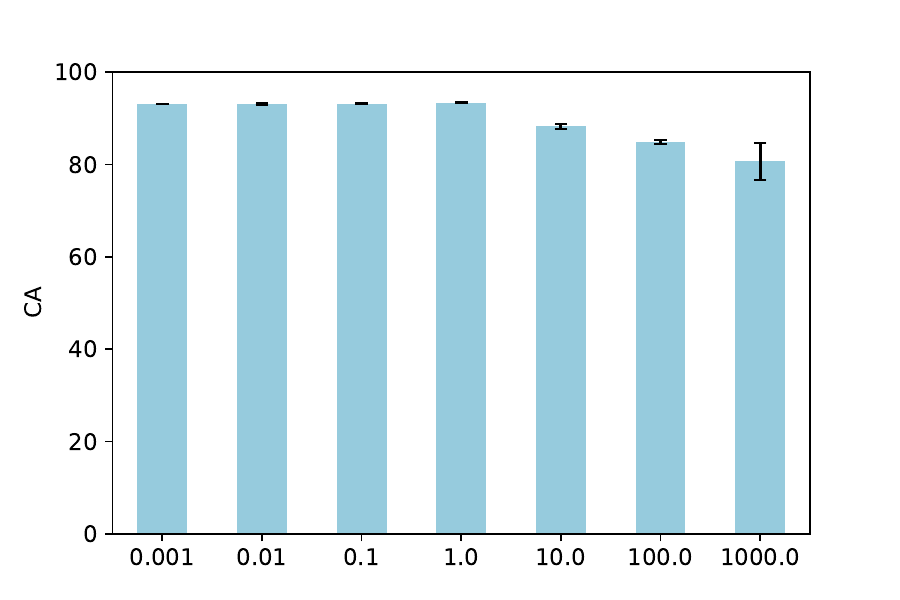}
    \end{subfigure}
    \begin{subfigure}{.24\textwidth}
      \centering
      \includegraphics[width=\textwidth]{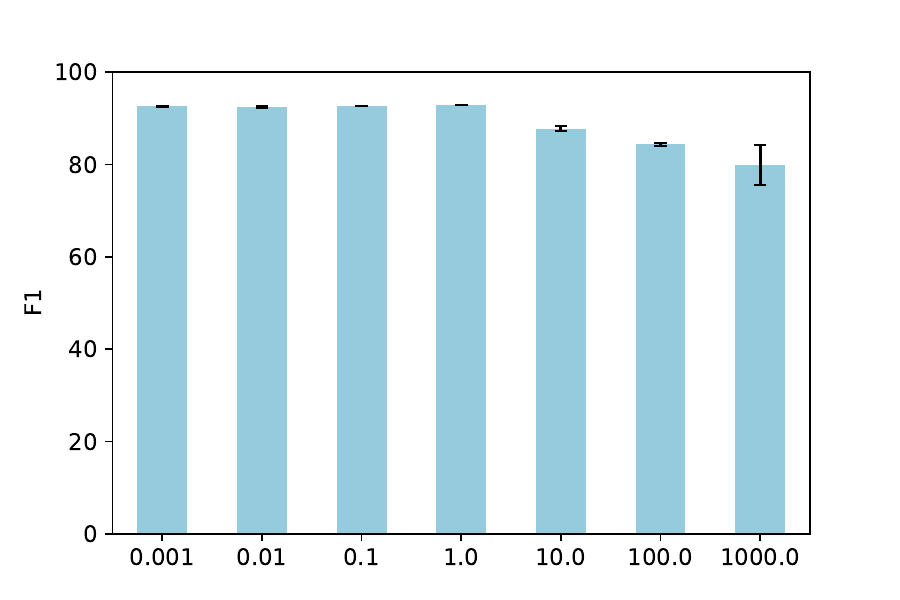}
    \end{subfigure}
    \begin{subfigure}{.24\textwidth}
      \centering
      \includegraphics[width=\textwidth]{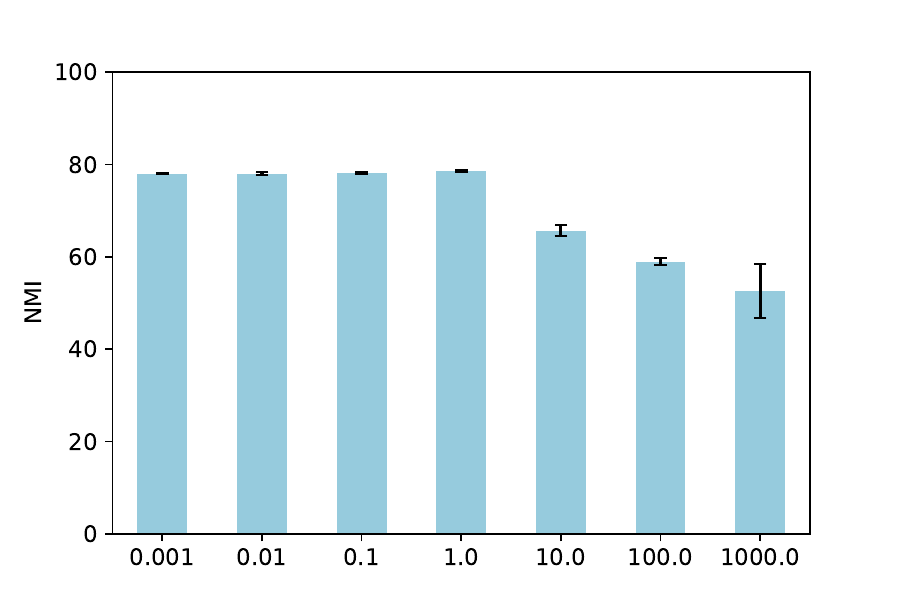}
    \end{subfigure}
    \begin{subfigure}{.24\textwidth}
      \centering
      \includegraphics[width=\textwidth]{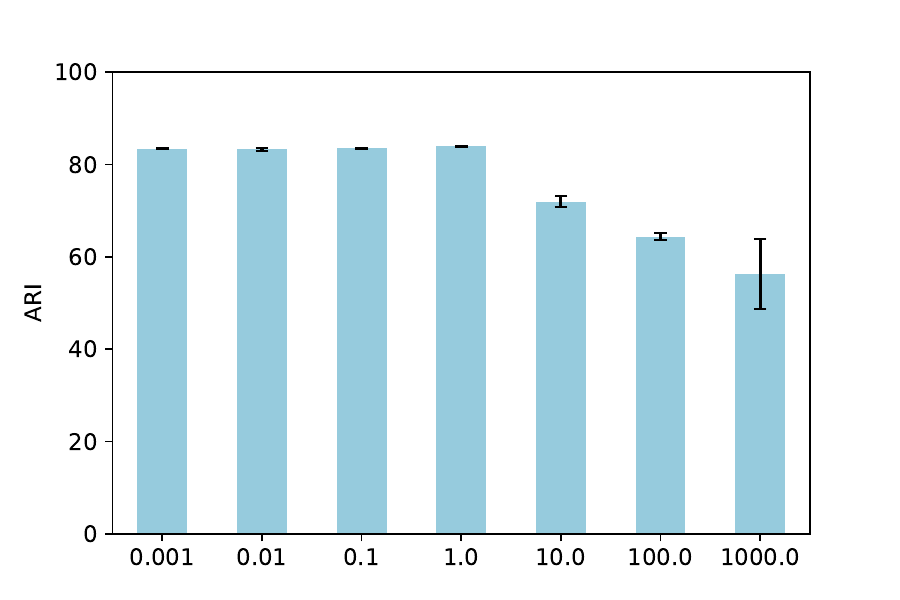}
    \end{subfigure}
    };
    \end{subfigure}
    \centering
    \begin{subfigure}{\textwidth}
    \centering
    \tikz\node[inner sep=0pt,
           label=west:\rotatebox{90}{\textBF{Photos}}]{\hspace{.2cm}
      \centering
      \begin{subfigure}{.24\textwidth}
      \centering
      \includegraphics[width=\textwidth]{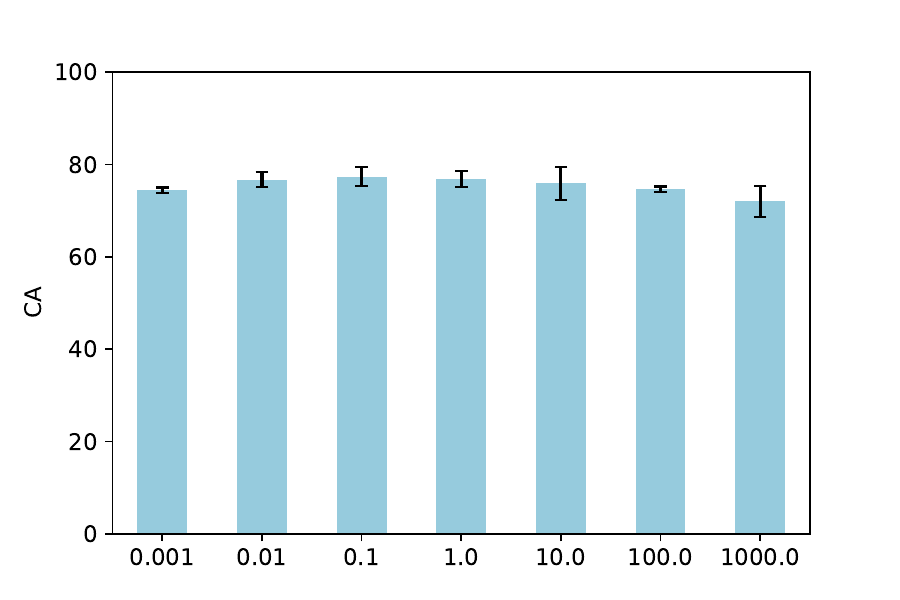}
    \end{subfigure}
    \begin{subfigure}{.24\textwidth}
      \centering
      \includegraphics[width=\textwidth]{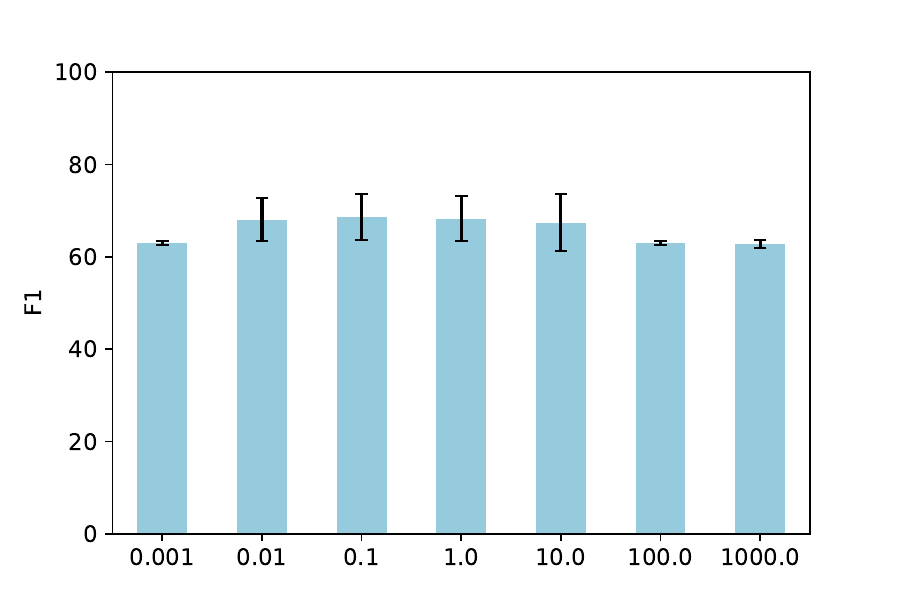}
    \end{subfigure}
    \begin{subfigure}{.24\textwidth}
      \centering
      \includegraphics[width=\textwidth]{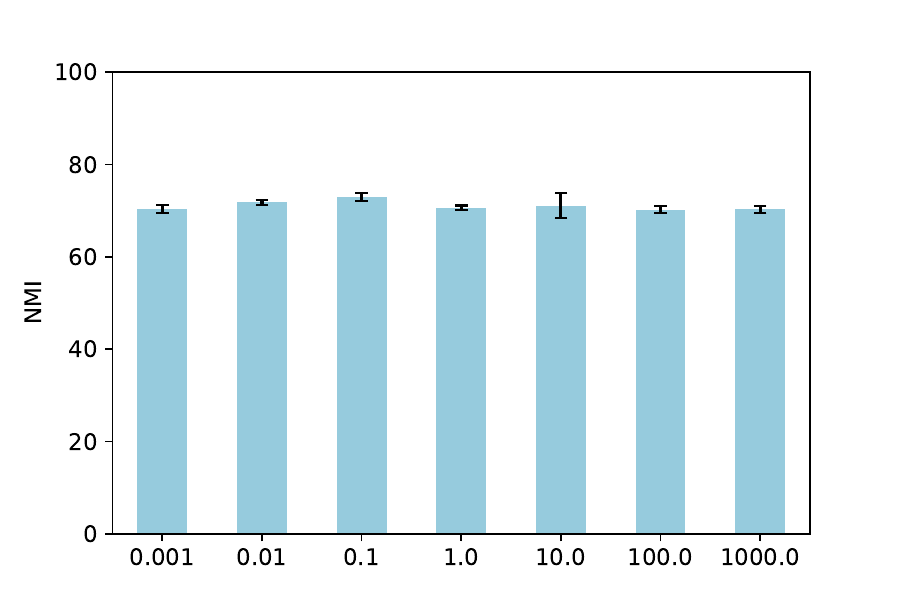}
    \end{subfigure}
    \begin{subfigure}{.24\textwidth}
      \centering
      \includegraphics[width=\textwidth]{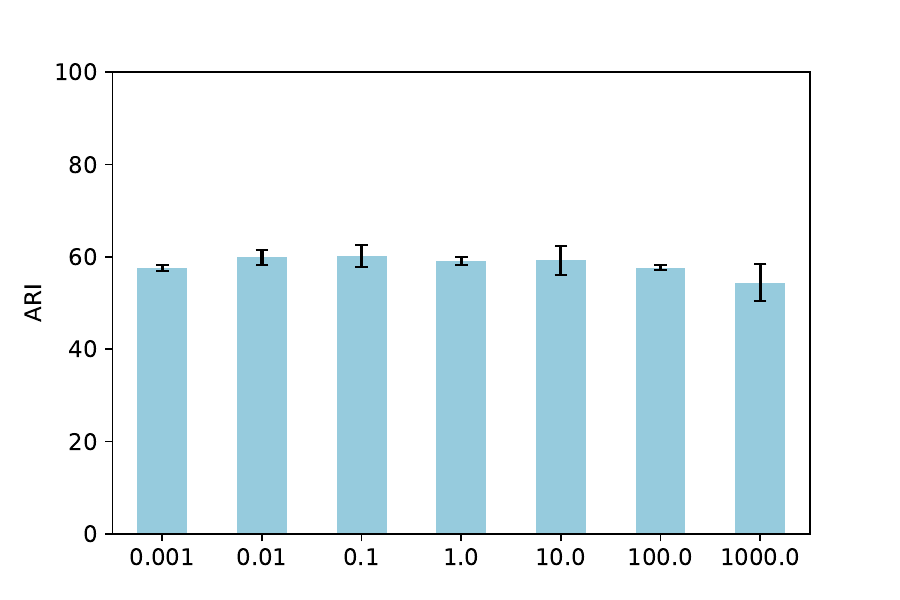}
    \end{subfigure}
    };
    \end{subfigure}
    \centering
    \begin{subfigure}{\textwidth}
    \centering
    \tikz\node[inner sep=0pt,
           label=west:\rotatebox{90}{\textBF{Computers}}]{\hspace{.2cm}
      \centering
      \begin{subfigure}{.24\textwidth}
      \centering
      \includegraphics[width=\textwidth]{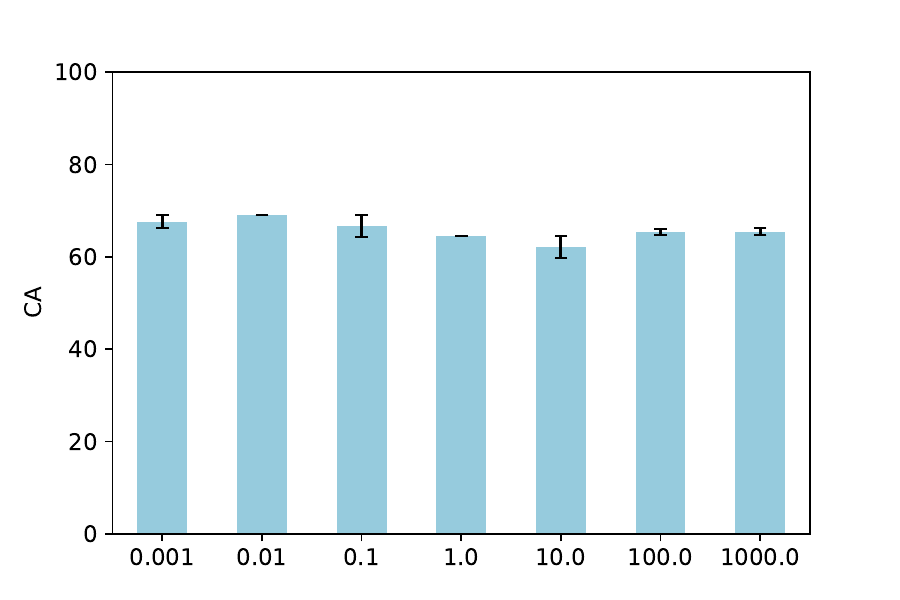}
    \end{subfigure}
    \begin{subfigure}{.24\textwidth}
      \centering
      \includegraphics[width=\textwidth]{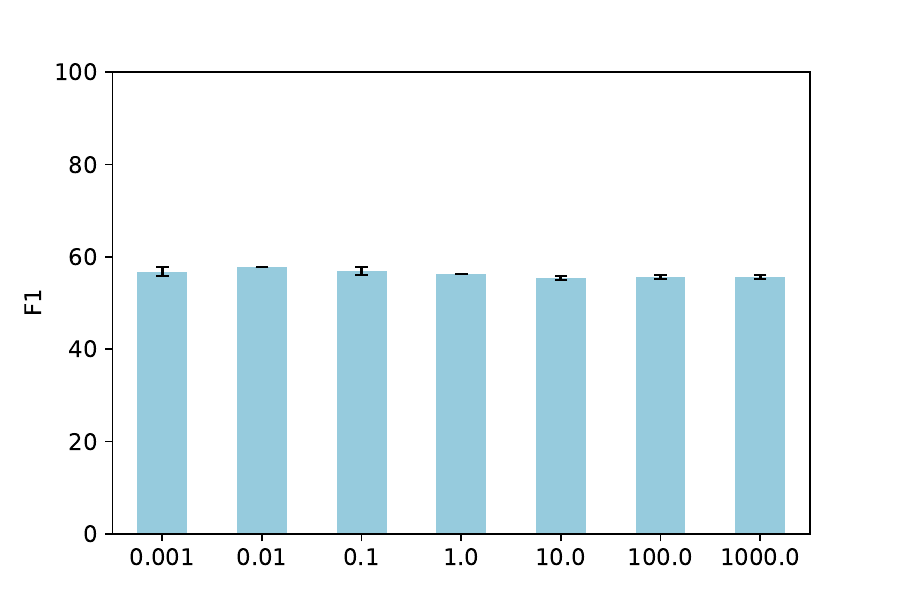}
    \end{subfigure}
    \begin{subfigure}{.24\textwidth}
      \centering
      \includegraphics[width=\textwidth]{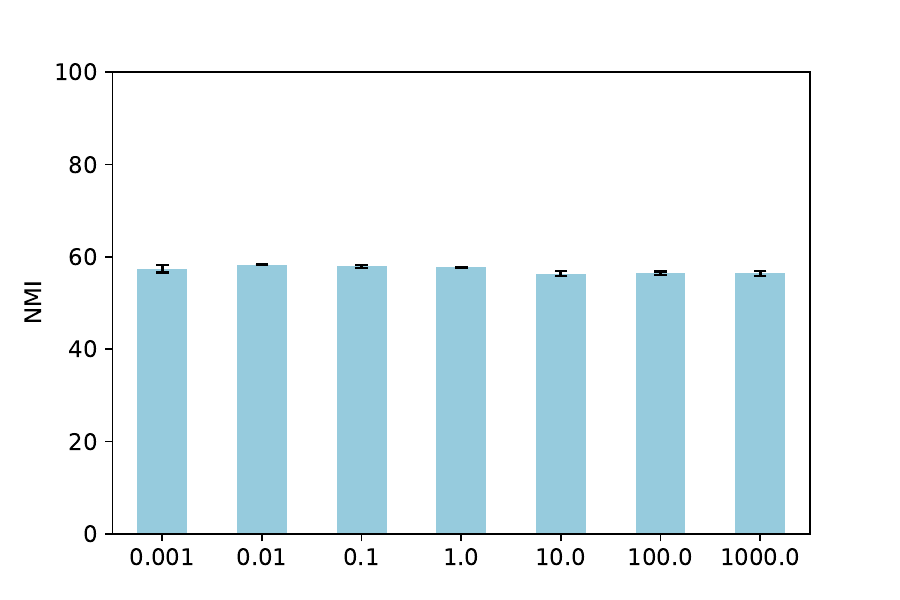}
    \end{subfigure}
    \begin{subfigure}{.24\textwidth}
      \centering
      \includegraphics[width=\textwidth]{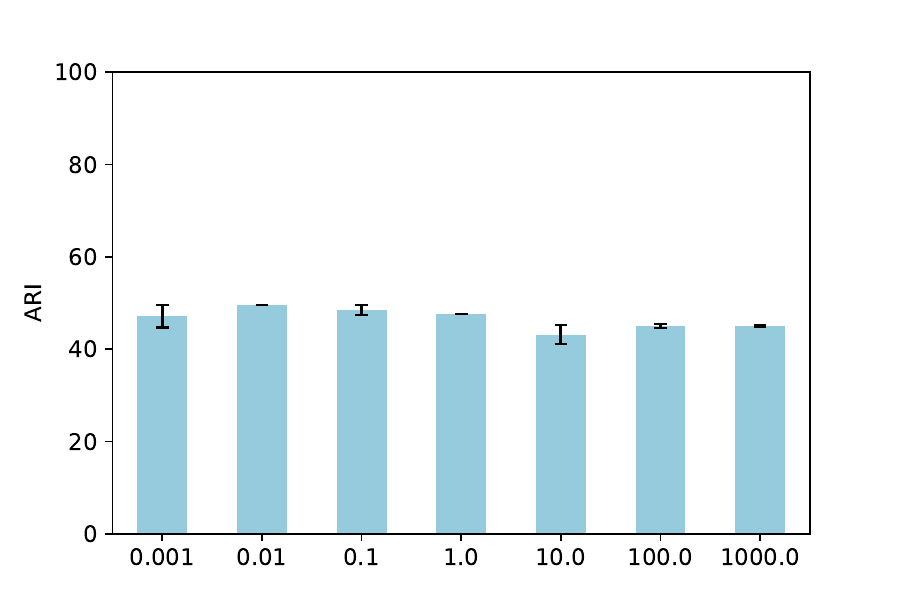}
    \end{subfigure}
    };
    \end{subfigure}
    \begin{subfigure}{\textwidth}
    \centering
    \tikz\node[inner sep=0pt,
           label=west:\rotatebox{90}{\textBF{ArXiV}}]{\hspace{.2cm}
      \centering
      \begin{subfigure}{.24\textwidth}
      \centering
      \includegraphics[width=\textwidth]{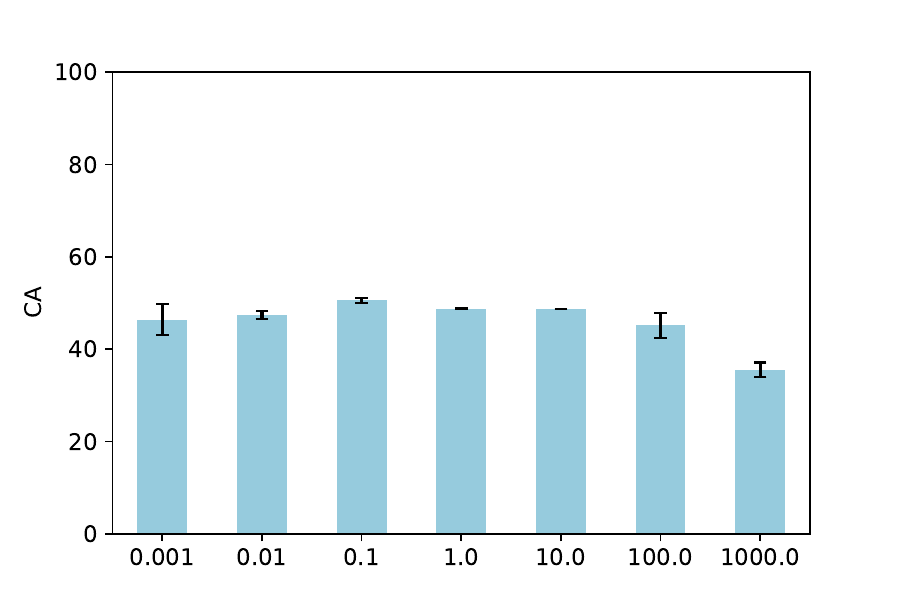}
    \end{subfigure}
    \begin{subfigure}{.24\textwidth}
      \centering
      \includegraphics[width=\textwidth]{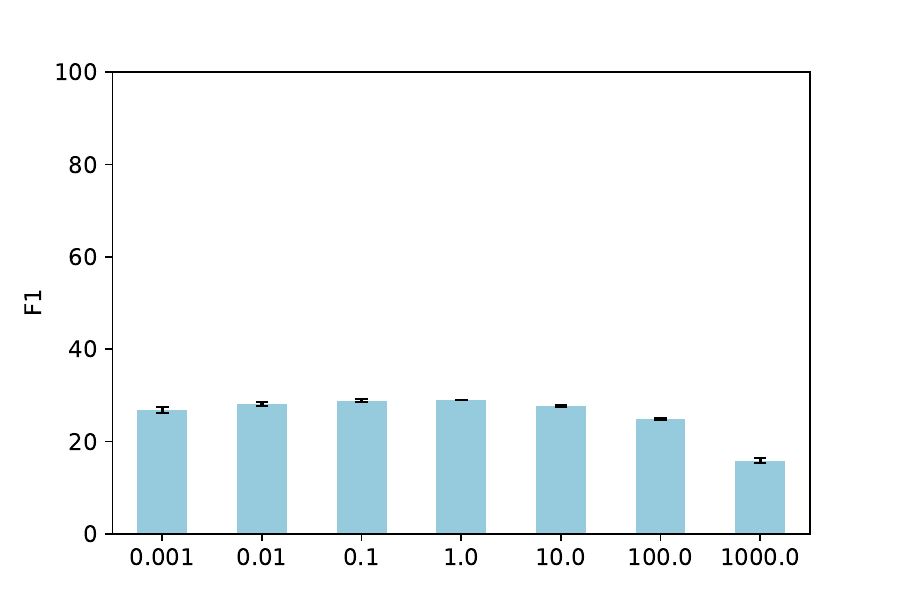}
    \end{subfigure}
    \begin{subfigure}{.24\textwidth}
      \centering
      \includegraphics[width=\textwidth]{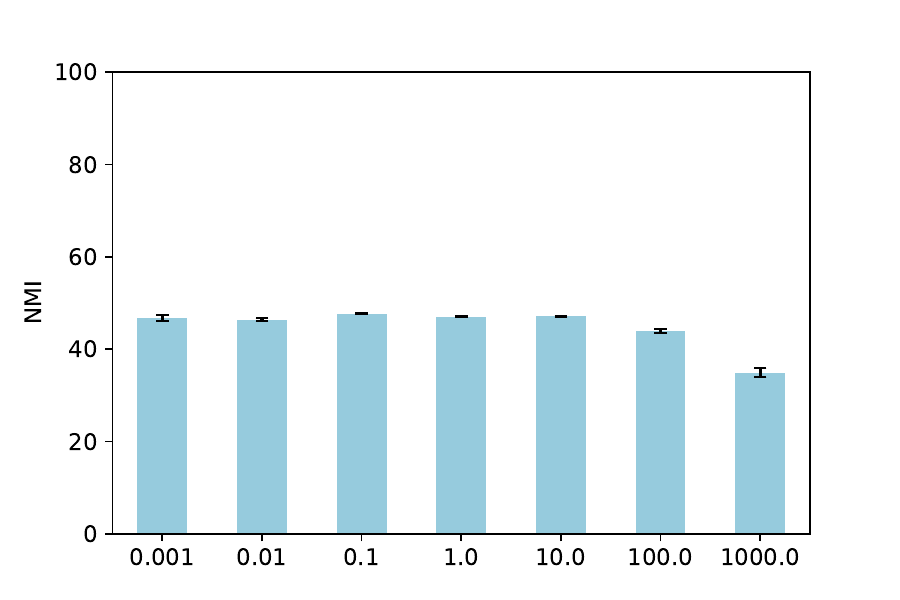}
    \end{subfigure}
    \begin{subfigure}{.24\textwidth}
      \centering
      \includegraphics[width=\textwidth]{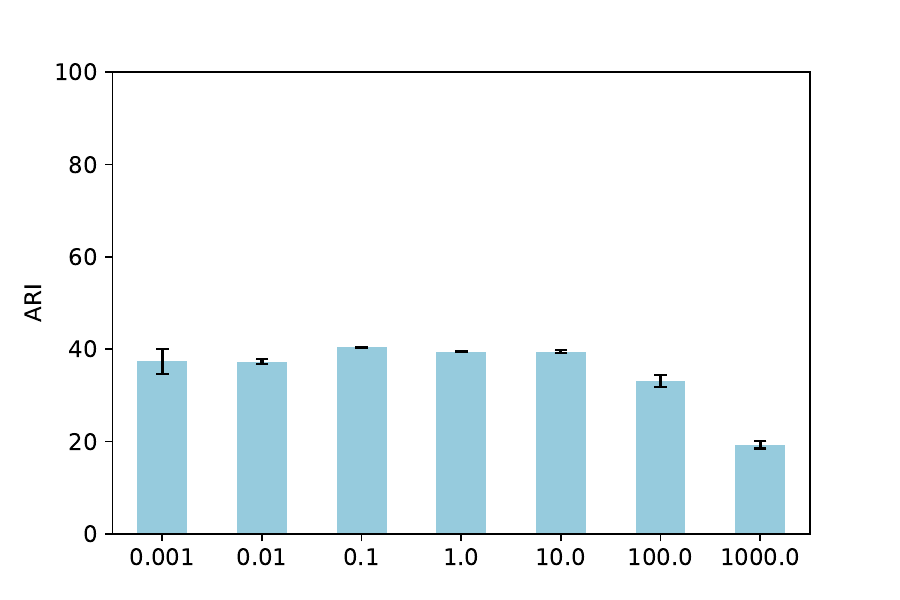}
    \end{subfigure}
    };
    \end{subfigure}
    \begin{subfigure}{\textwidth}
    \centering
    \tikz\node[inner sep=0pt,
           label=west:\rotatebox{90}{\textBF{Products}}]{\hspace{.2cm}
      \centering
      \begin{subfigure}{.24\textwidth}
      \centering
      \includegraphics[width=\textwidth]{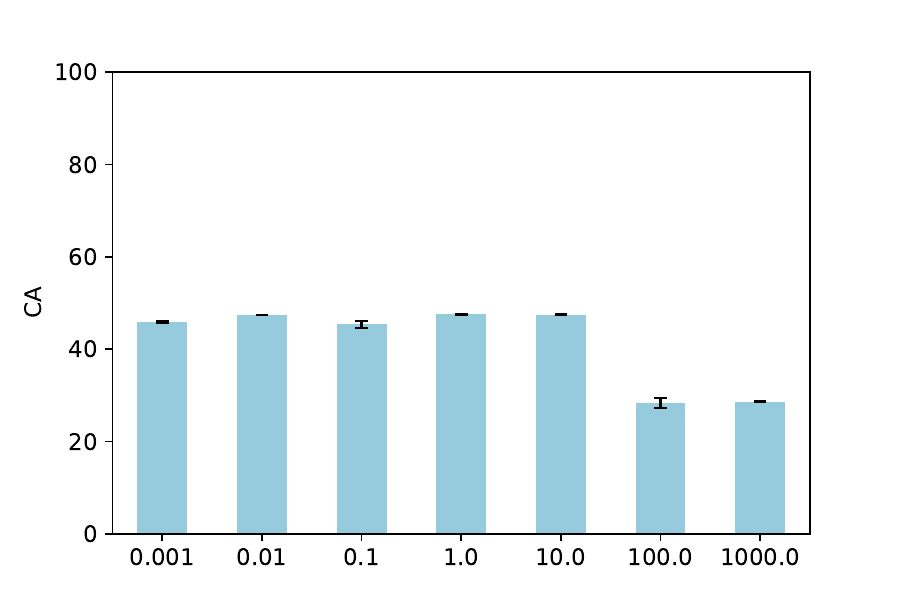}
    \end{subfigure}
    \begin{subfigure}{.24\textwidth}
      \centering
      \includegraphics[width=\textwidth]{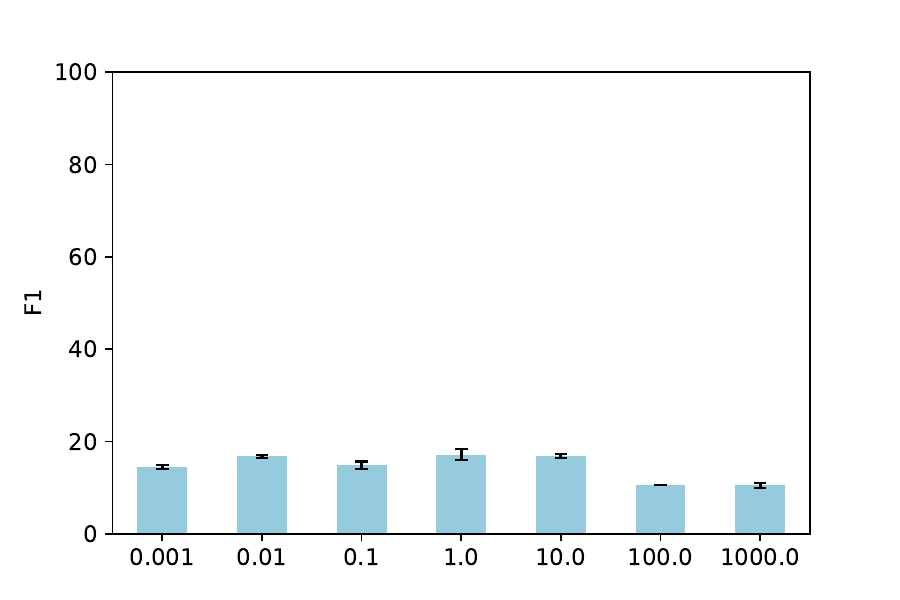}
    \end{subfigure}
    \begin{subfigure}{.24\textwidth}
      \centering
      \includegraphics[width=\textwidth]{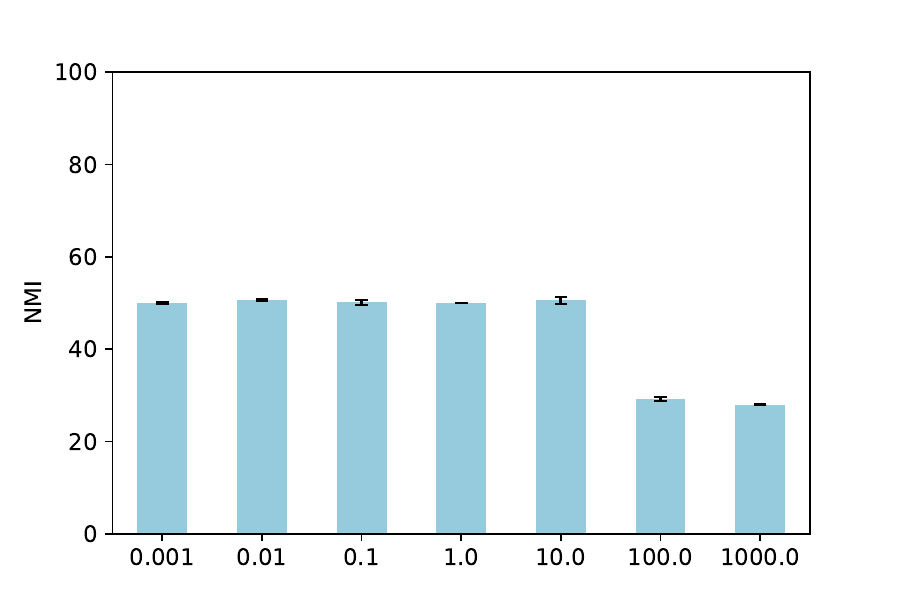}
    \end{subfigure}
    \begin{subfigure}{.24\textwidth}
      \centering
      \includegraphics[width=\textwidth]{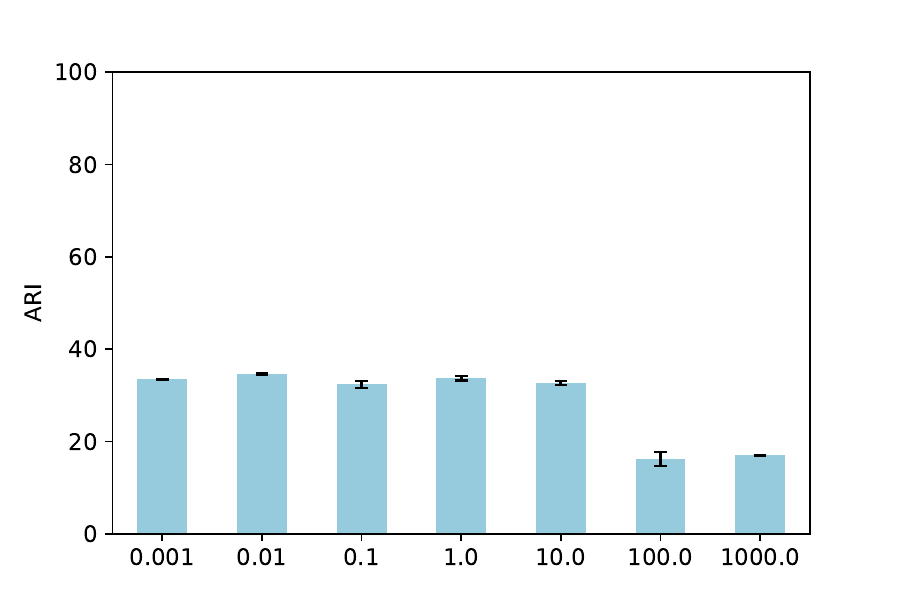}
    \end{subfigure}
    };
    \end{subfigure}\\
    {\setlength{\tabcolsep}{1.22cm}
    \begin{tabular}{cccc}
    \hspace{1.cm}\textBF{CA} & \textBF{CF1} & \textBF{NMI} & \textBF{ARI}\\
    \end{tabular}}
\caption{\textcolor{black}{Sensitivity of MvSCK in terms of CA, CF1, NMI and ARI according to the temperature parameter $T$.}}
\label{fig:sensitivity}
\end{figure}

\subsubsection{Temperature Parameter $\ T$}
In our experiments we tried various values for the temperature hyperparameter. Figure \ref{fig:sensitivity} shows the performance of MvSCK for different temperature values in different data sets and for the different clustering metrics. We see that on most datasets the performance remains mostly constant, but that larger values deteriorate the performance. As such, we advise using $T\leq 1$. In our experiments, we set it to $T=0.1$ for all datasets.

\subsubsection{ Kernel Feature Map}
In Table \ref{tab:kernels} we report the results of our approach using different kernels, namely an exact quadratic kernel, Nystroem approximations of the radial basis function (RBF), and a sigmoid kernel. For the approximations we use $10\times k$ components, apart from on the Products dataset, which is too large and for which we use $100$ components only. The results show that using different kernels gives similar performances, except on Products, where the dimension of the approximations would appear to be insufficient, leading to a poor approximation of the feature maps; the same is true for the sigmoid kernel on ArXiv. Note that in the context of RBF and sigmoid we use approximations, given that their exact feature maps are infinite-dimensional.

\begin{table}[h]
        \caption{MvSCK results and running times (in seconds) using different kernels.}
    \label{tab:kernels}
    \centering
    \scriptsize{    
    \begin{tabular}{@{}lcccccccccccc@{}}
    \toprule
     & \multicolumn{2}{c}{\textBF{ACM}} & \multicolumn{2}{c}{\textBF{DBLP}} & \multicolumn{2}{c}{\textBF{Photos}} & \multicolumn{2}{c}{\textBF{Computers}}&\multicolumn{2}{c}{\textBF{ArXiv}} & \multicolumn{2}{c}{\textBF{Products}} \\\cmidrule(lr){2-3}\cmidrule(lr){4-5}\cmidrule(lr){6-7}\cmidrule(lr){8-9}\cmidrule(lr){10-11}\cmidrule(lr){12-13}
    Kernel &     CA  &    {Time}  & CA  &    {Time} & CA  &    {Time} & CA  &    {Time}& CA  &    {Time}& CA  &    {Time} \\
    \midrule
    {Quadratic} &  \textBF{93.2} &  \textBF{0.2} & \textBF{ 93.1} &  \textBF{0.2} &  75.7 &  \textBF{0.9} &  \textBF{66.3} &  \textBF{6.0} &  48.3 &  20.4 &  \textBF{47.1} &  239.3 \\
    {RBF}  &  \textBF{93.2} &  0.8 &  \textBF{93.1} &  2.0 &  \textBF{76.2} &  2.4 &  65.0 &  6.5 &  \textBF{48.7} &  \textBF{15.8} &  18.6 &  325.2 \\
    {Sigmoid}  &  \textBF{93.2} &  0.9 &  93.0 &  1.9 &  74.7 &  2.3 &  65.6 &  6.4 &  34.0 &  16.6 &  33.2 &  \textBF{231.3} \\
    \bottomrule
    \end{tabular}
    }
\end{table}

\subsection{Ablation relative to the View-Importance Parameter  \texorpdfstring{$\boldsymbol\lambda$}{λ} (RQ5)}

\begin{figure}[h]
    \centering
    \includegraphics[width=.65\columnwidth]{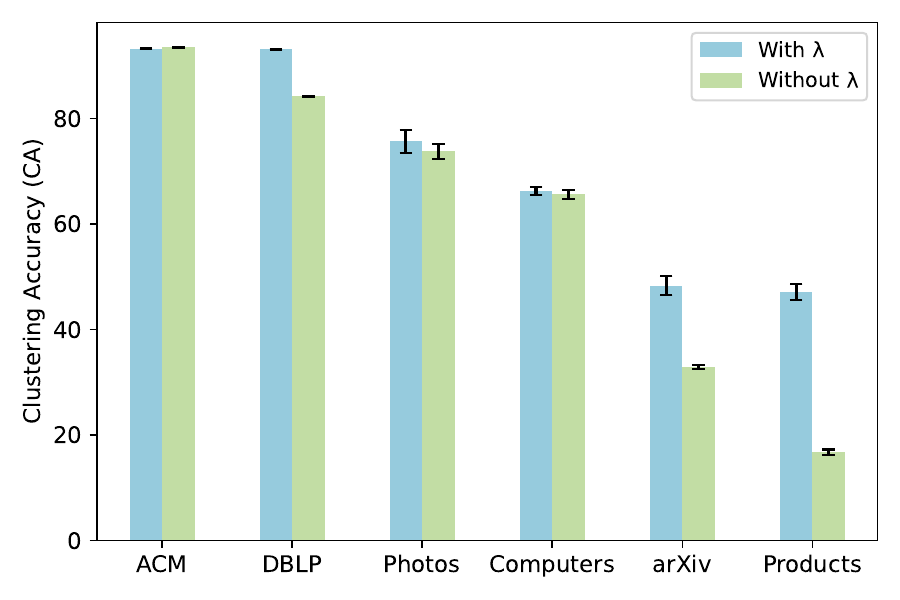}
    \caption{Clustering accuracy with and without the regularization vector $\boldsymbol{\lambda}$}
    \label{fig:lmbda}
\end{figure}

Figure \ref{fig:lmbda} shows the effect of the regularization vector $\boldsymbol{\lambda}=[\lambda_1,\ldots,\lambda_V]$ on the different datasets. $\boldsymbol{\lambda}$ is seen to boost performance on five out of the six datasets that we considered for ablation, and does not impair performance on the remaining dataset ACM.
Figure \ref{fig:lmbda} illustrates the impact of the regularization vector $\boldsymbol{\lambda}=[\lambda_1,\ldots,\lambda_V]$ on various datasets. The regularization vector $\boldsymbol{\lambda}$ improves performance in five of the six datasets analyzed during ablation studies. Notably, there is no deterioration in performance on the remaining ACM dataset, indicating that the regularization vector does not adversely affect the outcome in this particular case.

\section{Conclusion}
In conclusion, this paper introduced a novel scalable framework for multi-view subspace clustering, leveraging kernel feature maps in order to ensure an efficient computation of the consensus subspace affinity graph. In virtue of the properties intrinsic to kernel summation, the approach that we propose results in performance gains and significant speedup. Extensive experiments on real-world benchmark networks, whatever the degree of overlap, demonstrate the superiority of our algorithm over state-of-the-art methods, especially in the context of very large network datasets, where most other models used in our comparison failed to scale despite using the same computational resources.

\bibliography{sn-bibliography}

\end{document}